\documentclass[preprint,12pt]{elsarticle}

\usepackage{amssymb}
\usepackage{amsmath}

\usepackage{subcaption}
\usepackage{array}
\usepackage{changes}
\definechangesauthor[name={me}, color=blue]{me}
\usepackage{float}
\usepackage{xcolor}

\usepackage{longtable}
\usepackage{booktabs}
\makeatletter
\def\ps@pprintTitle{%
  \let\@oddhead\@empty
  \let\@evenhead\@empty
  \let\@oddfoot\@empty
  \let\@evenfoot\@empty
}
\makeatother

\begin{document}

\begin{frontmatter}

\title{Cross-Population White Matter Atlas Creation for Concurrent Mapping of Brain Connections in Neonates and Adults with Diffusion MRI Tractography}

\author[label1]{Wei Zhang\fnref{equal}}
\author[label1]{Yijie Li\fnref{equal}}
\author[label1]{Ruixi Zheng}
\author[label4]{Nir A. Sochen}
\author[label2]{Yuqian Chen}
\author[label2]{Leo R. Zekelman}
\author[label2]{Ofer Pasternak}
\author[label2,label3]{Jarrett Rushmore}
\author[label2]{Yogesh Rathi}
\author[label2]{Nikos Makris}
\author[label2]{Lauren J. O'Donnell}
\author[label1]{Fan Zhang\corref{cor1}}

\affiliation[label1]{organization={University of Electronic Science and Technology of China},
            city={Chengdu},
            country={China}}
\affiliation[label4]{organization={Tel-Aviv University},
            city={Tel Aviv},
            country={Israel}}
\affiliation[label2]{organization={Harvard Medical School},
            city={Boston},
            country={USA}}

\affiliation[label3]{organization={Boston University},
            city={Boston},
            country={USA}}


\fntext[equal]{These authors contributed equally to this work.}
\cortext[cor1]{Corresponding author: Fan Zhang, fan.zhang@uestc.edu.cn.}

\begin{abstract}
Comparing white matter (WM) connections between adults and neonates using diffusion MRI (dMRI) can enhance our understanding of typical brain development and help identify potential biomarkers for neurological disorders. Brain WM atlases are valuable tools for facilitating population-wise comparisons, allowing for the identification of WM differences between adult and neonatal brains. Currently, existing studies typically rely on atlases created for a specific population (either neonates or adults), where each atlas resides in its own space, preventing direct comparisons across populations. 
A cross-population WM atlas, created using both adult and neonatal imaging data, that allows for the concurrent mapping of brain connections in the two populations is still lacking. Therefore, in this study, we propose the neonatal/adult brain atlas, namely NABA, a WM tractography atlas that is built using dMRI data from both neonates and adults. NABA is created using a robust, data-driven tractography fiber clustering pipeline, allowing for group-wise WM atlasing across diverse populations, despite potential anatomical variability in the WM. This atlas serves as a standardized template for parcellating WM tractography in new individuals, enabling direct comparisons of WM tracts across the two populations. With the atlas, we perform a comparative study to examine the WM development of the neonates in comparison to the adult data. Our study encompasses four key analyses: (1) evaluating the feasibility of joint WM mapping between neonatal and adult brains using the atlas, (2) characterizing the development of the WM in different neonatal ages compared with adults, (3) assessing sex-related differences in neonatal WM development, and (4) investigating the effects of preterm birth on neonatal WM development. There are several key observations from our study. First, we demonstrate NABA’s advantages in identifying fiber tracts across both populations, with strong adaptability to age-specific anatomical variability. Furthermore, we observe rapid fractional anisotropy (FA) development in long-range association tracts, including the arcuate fasciculus and superior longitudinal fasciculus II, while intra-cerebellar tracts exhibit slower development in neonates. Third, neonatal females display overall faster FA development compared to males. Fourth, although preterm neonates exhibit a generally lower FA development rate than full-term infants, they show relatively higher FA growth rates in tracts, including the corticospinal tract, corona radiata-pontine pathway, and intracerebellar tracts. Overall, we show that the atlas can be a useful tool to investigate the WM development in neonates and adults.

\end{abstract}


\end{frontmatter}


\section{Introduction}
\label{Introduction}
The white matter (WM) enables information exchange and communication between different brain regions, playing a critical role in maintaining proper brain function in both health and disease. The development of WM is crucial because it enhances the efficiency and speed of neural communication, supporting cognitive growth, learning, motor coordination, and overall brain function throughout development \cite{Asato2010-xs,Lebel2018-ad}. The developmental effects of WM occur at vastly different time scales, in which the neonatal stage is one of the important periods with the most dramatic and substantial changes \cite{Dubois2014-my,Huang2006-yd,Sket2019-sx,Warrington2022-dp}. In comparison, the WM demonstrates relative stability across young adulthood, providing a valuable contrast for investigating early developmental WM changes in neonates. Comparison of the WM connections between adult and neonate brains can contribute to our understanding of normal brain development and assist identification of potential biomarkers for neurological disorders \cite{Zhai2003-ma}.

Diffusion MRI (dMRI) is an advanced MRI technique that can probe the diffusion of water molecules in biological tissues to characterize the underlying microstructure \cite{Basser1994-mh}. dMRI enables a computational process, namely tractography, that uniquely enables in-vivo reconstruction of the brain’s WM connections at macro scale \cite{Basser2000-gg,Zhang2022-he}. dMRI tractography has become a popular tool for studying the development of the brain's WM connections at early ages \cite{Pannek2014-ue,Verschuur2024-gq}. For example, Wilkinson et al. \cite{Wilkinson2017-yw} and Tak el al. \cite{Tak2016-uw} investigated the arcuate fasciculus (AF) using dMRI tractography and found hemispheric asymmetry along with moderate increases in fractional anisotropy (FA) with age. Liu et al. \cite{Liang2022-el} examined the superior longitudinal fasciculus (SLF) I to III tracts using tractography and found that SLF II is the least mature. 

To provide a reference and evaluate WM tract development, the comparison of fiber tracts using tractography at different ages is important. However, direct comparison of WM between adults and neonates presents significant challenges due to substantial differences in brain size and axonal myelination \cite{Ouyang2019-vj}. While spatial normalization of brain images to a common space is a standard approach for such comparisons, previous neonatal studies \cite{Cabral2022-cz,Truzzi2023-np}, \cite{Gaillard2001-pq,Kazemi2007-lp} have identified two major challenges in this methodology. First, the neonatal scalp is notably thinner and lies in closer proximity to the cortex compared to adults, which can introduce misalignment errors when registering adult imaging data to neonatal data. Second, the neonatal cerebellum exhibits a significantly smaller volume than its adult counterpart, further complicating cross-age registration due to scale and structural differences. As a result, traditional image registration between adult and neonate brains into a common space introduces additional errors. One recent study mapped neonates, adults, and macaques to a common space to compare cortical connectivity \cite{Warrington2022-dp}, using white matter tractography to identify corresponding cortical regions. This study showed the potential of using white matter fiber tracts to build the correspondence between neonates and adults for direct brain comparison. 

Brain atlases serve as essential tools for standardized representation of the anatomical structures and functional regions within the brain \cite{Collins1994-vo,Laird2009-hk,Mazziotta2001-tl} to perform comparisons of brains at different ages \cite{Zhang2018-lx}. In general, a brain atlas provides a common template derived from a group of individuals, which can be applied to diverse individuals to identify subject-specific anatomically segregated brain regions. In the dMRI tractography literature, many WM atlases have been proposed to enable the subject-specific identification of anatomical WM connections \cite{Guevara2012-id,Roman2017-hd,Yeh2018-cx,Zhang2018-lx}. Nevertheless, existing WM atlases primarily originate from data collected from adults. While research has been performed by applying adult atlases to neonatal brains for population-wise comparison \cite{Warrington2022-dp,Zhang2018-lx}, it might not capture the large inter-population WM anatomical variability between neonates and adults. For example, the ORG atlas \cite{Zhang2018-lx}, created for the adult population, provides consistent automated white matter tract parcellation across the lifespan. However, while it can be applied to neonatal populations, it may fail to capture the subtle inter-population anatomical variability and the unique white matter development of neonates. Furthermore, studies \cite{Dimitrova2021-hg,Makropoulos2016-oc,Warrington2022-dp} examining the brains across different ages, focusing on the cortical surface and brain morphology, have highlighted the necessity of creating atlases specific to neonates for accurate anatomy representation. Although using separate atlases for neonates and adults ensures anatomical alignment within each population (e.g., aligning the AF in neonates to the same tract in adults), it cannot guarantee alignment in a shared mapping space where fiber bundles with similar spatial trajectories are aligned across populations. To the best of our knowledge, a white matter atlas spanning both neonatal and adult populations to analyze whole-brain WM differences has not yet been established. Thus, there is a high need for a cross-population WM atlas that enables the concurrent mapping of WM fiber tracts between neonates and adults. Such an atlas can help bridge the current gap in tract-specific developmental research, providing a unified framework for quantifying and comparing fine-grained, tract-wise developmental trajectories across populations.

In this study, we create the neonatal/adult brain atlas, namely NABA, a WM tractography atlas that is built using dMRI data from both neonates and adults. The NABA atlas is created using a robust, data-driven tractography fiber clustering pipeline \cite{Li2024-kr,ODonnell2012-kp,ODonnell2007-zx,Zhang2018-lx,Zhang2020-nh}, allowing for group-wise WM atlasing across diverse populations, despite potential anatomical variability in the WM. We leverage two large-scale dMRI datasets, i.e., the Developing Human Connectome Project (dHCP) \cite{Makropoulos2018-vj} acquired from neonates and the Human Connectome Project Young Adult (HCP-YA) \cite{Elam2021-gc} from young adults. Using the newly created atlas, we examine the development of WM fiber tracts in neonatal brains compared to those of healthy adults. Our study encompasses four key analyses: (1) evaluating the feasibility of concurrent WM mapping between neonatal and adult brains using the atlas, (2) characterizing the development of WM in neonates compared with adults, (3) assessing sex-related differences in neonatal WM development, and (4) investigating the effects of preterm birth on neonatal WM development.

\section{Material and Methods}
\label{Material and methods}
\begin{figure}[htbpp]
  \centering
  \includegraphics[width=1\textwidth]{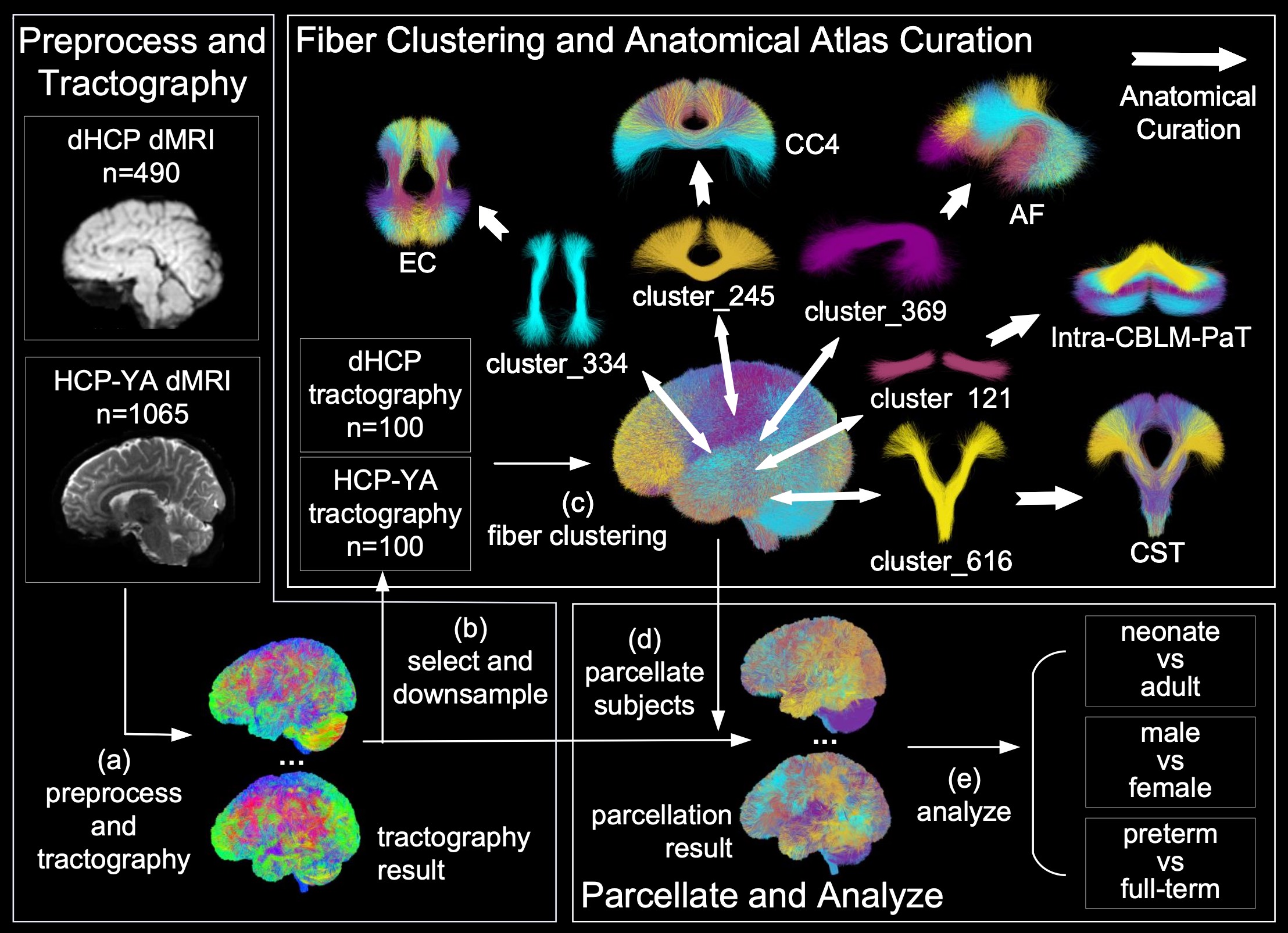}
    \caption{Method Overview. (a) First, the dMRI data of 490 dHCP and 1065 HCP-YA subjects are preprocessed, and tractography is performed. Next, the anatomical atlas is constructed. (b) A total of 100 subjects with balanced age and sex are selected from each dataset. Their tractography is downsampled to the same level. (c) Their tractography data is combined and processed using fiber clustering. Anatomical labels are then assigned to each fiber cluster to define white matter tracts. Different clusters are color-coded for visualization, and those having same anatomical label are grouped together to anatomical tracts. (d) Then, subject-specific tract identification is performed for all individuals using the constructed atlas. Finally, statistical analysis of white matter tracts is conducted across different subject groups.}
  \label{overview}
\end{figure}
Figure \ref{overview} provides an overall conceptual representation of the methodology used to create the joint NABA atlas and the data analysis to examine WM development in neonates and adults. In the rest of the paper, we first describe the dMRI datasets and preprocessing steps used in our study (Section \ref{2.1}), followed by the creation of the WM atlas (Section \ref{2.2}). Then, we introduce the process of applying the created atlas to each neonate subject and adult subject to identify the subject-specific fiber tract and subsequently the WM metrics extracted from each fiber tract (Section \ref{2.3}). Finally, we describe the statistical analyses used to study the development of WM tracts in neonates in comparison to adults (Section \ref{2.4}). 

\subsection{Participants, dMRI Dataset, and Data Processing}
\label{2.1}
This study utilizes dMRI datasets from two sources: the dHCP data obtained from neonates \cite{Makropoulos2018-vj} and the HCP-YA data obtained from young healthy adults \cite{Elam2021-gc}. In brief, the neonatal participants are drawn from the second release of the dHCP study. We exclude the individuals from the entire cohort whose dMRI scans have a radiology assessment score greater than 3, indicating larger or more clinically significant incidental findings, according to \cite{Warrington2022-dp}. In total, 490 neonatal participants are used, including 213 females and 277 males born at 24.57 to 42.29 weeks postmenstrual age (PMA) and scanned at 29.29 to 45.14 weeks PMA. The adult data are drawn from the HCP-YA study. We use all 1065 individuals with available dMRI scans, including 575 females and 490 males scanned at 22 to 37 years old.

The dMRI data of the above participants are publicly available in the HCP-YA (db.humanconnectome.org) and dHCP (www.developingconnecto\\me.org) databases. The detailed acquisition parameters are described in  \cite{Elam2021-gc} and  \cite{Makropoulos2018-vj}. In brief, the dHCP data was acquired on a 3T Philips Achieva MR scanner with TE = 90 ms, TR = 3800 ms, voxel size $= 1.5 \times 1.5 \times 1.5 \text{mm}^3 $, and a total of 300 volumes including 20 baseline images and 280 diffusion-weighted images at b = 400/1000/2600 $ \text{s/mm}^3 $. The HCP-YA data was acquired on a customized 3T Connectome Siemens Skyra scanner with TE = 89.5 ms, TR = 5520 ms, voxel size $= 1.25 \times 1.25 \times 1.25 \text{mm}^3 $, and a total of 288 volumes, including 18 baseline images and 90 diffusion-weighted images at b = 1000/2000/3000 $ \text{s/mm}^3 $. For both datasets, we use the dMRI data already preprocessed and provided in the public databases.

Tractography is performed to reconstruct the whole brain WM pathways using the two-tensor Unscented Kalman Filter (UKF) algorithm \cite{Farquharson2013-xc,Vos2013-vn}. This method accounts for crossing fibers and offers sensitive and reliable fiber tracking for consistent fiber tracking across a wide range of populations, ages, and scan acquisitions \cite{Li2024-kr,Xue2023-km,Zhang2020-qj,Zhang2018-lx}. Tractography parameters are set the same as those used in our previous work \cite{Li2024-kr,Xue2023-km,Zhang2020-qj,Zhang2018-lx}. The diffusion tensors and their microstructure measure FA values at each fiber point location are calculated during tractography. Additionally, the dHCP tractography is uniformly enlarged by 1.5× using a hardened affine transform based on a predefined scaling matrix, so that its overall brain size matched the spatial level of HCP-YA.

\subsection{Cross-Population Fiber Clustering Atlas Creation}
\label{2.2}
Following the computation of tractography data for each subject, a whole-brain fiber clustering atlas is constructed using a robust, data-driven pipeline  \cite{ODonnell2012-kp,ODonnell2007-zx,Zhang2018-lx}, available in the whitematteranalysis (WMA) software\footnote{https://github.com/SlicerDMRI/whitematteranalysis} via Slicer\-DMRI \cite{zhang2020slicerdmri,norton2017slicerdmri}. This WMA pipeline has previously demonstrated success in generating tractography atlases for both white matter \cite{Li2024-kr,Zhang2018-lx} and cranial nerves \cite{Zeng_undated-iw,Zhang2020-nh}. This atlas is fundamentally composed of sets of streamlines derived from multiple subjects that cluster together to form labeled anatomical structures. The process involves two major steps: first, a groupwise registration aligns tractography data from all subjects into a common space; second, spectral clustering is applied to divide the aligned tractography into distinct fiber clusters, as described in detail below.

First, for each of the dHCP and HCP-YA datasets, we select 100 subjects (with a 50\% male-female ratio and no significant age difference, $p<0.05$) for atlas creation. To enable efficient computation, we downsample the number of WM streamlines to $21\times10^4$ per subject, thus leading to a total of $21\times10^4 \times 200 = 42$ million streamlines for the atlas creation. Second, a group-wise registration of the downsampled tractography data across all 200 subjects is performed to align all data into the same space \cite{ODonnell2012-kp}. The group-wise tractography registration method performs an entropy-based registration in a multiscale, coarse-to-fine manner based on the pairwise fiber trajectory similarity (mean closest point distance) across subjects, which can effectively handle the different brain sizes between the neonates and adults. Third, spectral clustering is employed to generate a high-dimensional fiber clustering atlas \cite{ODonnell2007-zx}, dividing the aligned tractography into $K$ clusters, where $K$ is a user-specified parameter that determines the parcellation scale. We chose $K = 800$, which is shown to be a good parcellation scale of the whole brain WM in previous studies \cite{ODonnell2017-fy,Wu2021-rf,Zhang2018-lx}. The spectral embedding creates a space that robustly represents each fiber based on its affinity to all other fibers across subjects. This representation provides a robust feature vector, or "fingerprint," describing each fiber for clustering. A key advantage of this fiber similarity matching is its robustness to local fiber tract variations, enhancing morphological consistency across subjects despite potential anatomical differences between adult and neonatal brains. Fourth, we perform anatomical curation of the fiber clusters by annotating each cluster with an anatomical label belonging to a certain anatomical tract (e.g., the corticospinal tract). To achieve this, we utilize the ORG atlas, which was constructed using HCP-YA data, as a reference. The two atlases are aligned through tractography-based registration \cite{ODonnell2012-kp}, and the mean closest point distances \cite{Moberts2005-yk,ODonnell2007-zx} are calculated to determine the spatial correspondence between clusters in the new atlas and those in the ORG atlas. Each cluster in the new atlas is then labeled according to the nearest corresponding cluster in the ORG atlas. In total, our NABA atlas contains an anatomical tract parcellation including 78 major anatomical WM tracts (Table \ref{table1}). The NABA atlas and the user documentation are available at https://github.com/azurezhangwei/NABA-Atlas.

\begin{table}[htbp]
\centering
\caption{The 78 tracts included in our atlas. Except for tracts in commissural bundles and MCP, other tracts have left and right parts (L, R).}
\label{table1}

\renewcommand{\arraystretch}{1.1}
\begin{tabular}{
>{\centering\arraybackslash}p{3.6cm}
>{\raggedright\arraybackslash}p{9.1cm}}
\toprule
\textbf{Category} & \multicolumn{1}{c}{\textbf{Tract name(s)}} \\
\midrule
Association tracts & AF(L,R), EC(L,R), EmC(L,R), ILF(L,R), IOFF(L,R), MdLF(L,R), SLF-I(L,R), SLF-II(L,R), SLF-III(L,R), UF(L,R) \\
Commissural tracts & CC1, CC2, CC3, CC4, CC5, CC6, CC7 \\
Limbic tracts & CB-D(L,R), CB-V(L,R) \\
Projection tracts & CST(L,R), CR-F(L,R), CR-P(L,R), SF(L,R), SO(L,R), SP(L,R), TF(L,R), TO(L,R), TT(L,R), TP(L,R) \\
Cerebellar tracts & CPC(L,R), ICP(L,R), Intra-CBLM-I\&P(L,R), Intra-CBLM-PaT(L,R), MCP, SCP(L,R) \\
Superficial tracts & Sup-F(L,R), Sup-FP(L,R), Sup-O(L,R), Sup-OT(L,R), Sup-P(L,R), Sup-PO(L,R), Sup-PT(L,R), Sup-T(L,R) \\
\bottomrule
\end{tabular}

\vspace{0.3cm}
\small
\begin{tabular}{@{}p{\textwidth}@{}}
\textit{Tract name abbreviation}: AF, arcuate fasciculus; EC, external capsule; EmC, extreme capsule; ILF, inferior longitudinal fasciculus; IOFF, inferior occipito-frontal fasciculus; MdLF, middle longitudinal fasciculus; SLF-I, superior longitudinal fasciculus I; SLF-II, superior longitudinal fasciculus II; SLF-III, superior longitudinal fasciculus III; UF, uncinate fasciculus; CC1–CC7, corpus callosum 1–7; CB-D, cingulum bundle dorsal; CB-V, cingulum bundle ventral; CST, corticospinal tract; CR-F, corona radiata frontal; CR-P, corona radiata parietal; SF, striato-frontal; SO, striato-occipital; SP, striato-parietal; TF, thalamo-frontal; TO, thalamo-occipital; TT, thalamo-temporal; TP, thalamo-parietal; CPC, cortico-ponto-cerebellar; ICP, inferior cerebellar peduncle; Intra-CBLM-I\&P, intracerebellar input and Purkinje tract; Intra-CBLM-PaT, intracerebellar parallel tract; MCP, middle cerebellar peduncle; SCP, superior cerebellar peduncle; Sup-F, superficial-frontal; Sup-FP, superficial-frontal-parietal; Sup-O, superficial-occipital; Sup-OT, superficial-occipital-temporal; Sup-P, superficial-parietal; Sup-PO, superficial-parietal-occipital; Sup-PT, superficial-parietal-temporal; Sup-T, superficial-temporal.
\end{tabular}
\end{table}

\subsection{Subject-specific Tract Identification and Diffusion Measure Extraction}
\label{2.3}
For each neonate and adult brain under study, subject-specific tractography parcellation is performed by applying the curated atlas to the whole brain tractography to identify the anatomical fiber tracts. In detail, the whole brain tractography of each subject is aligned into the atlas space using the tractography-based registration \cite{ODonnell2012-kp}. The aligned subject-specific tractography data is spectrally embedded into the atlas space by computing the spatial distances of its streamlines to the atlas, followed by the assignment of each fiber to the closest atlas cluster \cite{ODonnell2007-zx}. Anatomical tract identification of each subject is conducted by finding the subject-specific clusters that correspond to the annotated atlas clusters. 

Diffusion measures are extracted from each parcellated anatomical tract to quantify the microstructural properties of the brain white matter connections. In this study, we focus on FA and the number of streamlines (NoS) measures, which are sensitive to the changes in the underlying cellular microstructure in brain tissues \cite{Zhang2022-he}. FA is a widely used quantitative microstructure measure to reflect the diffusion anisotropy of the water molecules. NoS is a dMRI measure that estimates the white matter connection strength. 

\subsection{Statistical Analysis}
\label{2.4}
We conduct the following analyses of WM fiber tracts to study WM development using the NABA atlas. First, we evaluate the applicability and robustness of the NABA atlas for concurrent WM mapping in neonatal and adult brains. Next, we investigate the development of WM in different neonatal ages in comparison to adults. Finally, we perform two group-wise comparisons within the neonatal cohort—male vs. female and preterm vs. full-term—to examine the impact of sex and premature birth on the development of WM tracts, respectively.

\subsubsection{Concurrent WM Mapping between Neonate and Adult Brains}
A key feature of the NABA atlas is its ability to leverage both neonatal and adult brain data, enabling the identification of common fiber tracts for concurrent WM mapping across these two populations. To assess the applicability and robustness of this concurrent WM mapping, we compare the NABA atlas with the ORG atlas \cite{Zhang2018-lx}, which was created exclusively using adult data. For quantitative evaluation, we compute the identification rate (IR) for each fiber tract in both the dHCP and HCP-YA datasets using the NABA and ORG atlases. Specifically, for each fiber tract, we obtain two tract identification results per subject: one identified using the NABA atlas and the other using the ORG atlas. A fiber tract is considered successfully identified if it contains at least 10 streamlines, and its IR is defined as the percentage of successfully identified tracts across all subjects in a population (n=490 dHCP neonates or n=1065 HCP-YA adults). This threshold is chosen because previous studies have commonly used 1–20 streamlines to define the presence of a tract \cite{Zhang2018-lx,brown2011brain,drakesmith2015overcoming}, and 10 serves as a moderate and robust choice considering the lower streamline density in neonatal data. With the computed IRs, we test whether the tract identification results using the NABA atlas show a significant improvement compared to the ORG atlas. To do so, for each of the dHCP and HCP-YA datasets, we conduct a paired t-test on the IRs of all fiber tracts between the two atlases.

\subsubsection{Developmental Changes of WM in Neonates and Adults}
The NABA atlas provides a common space for both neonatal and adult brains, allowing for direct WM comparison between the populations despite the large anatomical variability. This serves as a valuable tool for using adult brains as a reference to explore early developmental trends in WM in different neonatal ages. In our study, using our NABA atlas, we examine the developmental changes of WM microstructure in neonates from the dHCP dataset (scanned at  29.29-45.14 weeks PMA) with respect to the HCP-YA adult brains (scanned at 22–37 years). For the dHCP data, to eliminate any potential effects associated with preterm birth, we analyze only the full-term neonates (n=307, age at birth $\geq$ 32 weeks PMA), as the threshold of 32 weeks PMA is commonly used to define very preterm infants in neonatal research\cite{ohuma2023national,perin2022global,Warrington2022-dp}. For the HCP-YA, we include all adult subjects (n=1065). To quantify the developmental changes in each fiber tract, we calculate how the tract-averaged FA measure varies with age in each population. For dHCP, following the analysis by Warrington et al. \cite{Warrington2022-dp}, a general linear model (GLM) is applied to assess the relationship between the FA and gestational age in weeks at the time of the scan. The resulting beta coefficients are used to quantify the rate of change with age, and the p-values (Bonferroni corrected among all tracts) are used to determine whether there are statistically significant developmental changes. We fit a similar linear model for HCP-YA, obtaining a consistent measure of beta coefficients and enabling direct comparison of developmental trajectories across populations. The obtained beta coefficients of each tract from the two datasets are compared to assess the WM development rate and the growth of FA values as age increases are plotted for visual comparison.

\subsubsection{Sex Effects in WM Development of Neonatal Brains}
Understanding the sex effects on WM development is crucial for comprehending how sex-specific factors influence early brain connectivity, developmental trajectories, and susceptibility to neurodevelopmental disorders \cite{Khan2024-rq}. In our study, using our NABA atlas, we explore sex differences in the developmental changes of WM in neonates. To eliminate any potential effects associated with pre-term birth, we include only the full-term neonates (n=307, age at birth $>=$ 32 weeks PMA). Furthermore, to eliminate age differences, we match males and females by age, selecting the largest set of age-matched pairs, including 116 female and 116 male neonates. Then, within each group, for each tract, we perform a GLM analysis between the subject’s tract FA and gestational age to quantify the rate of WM change with age (i.e., the beta coefficient).

\subsubsection{Preterm Birth Effects in WM Development of Neonatal Brains}
Premature birth is widely recognized as a crucial factor that causes profound disruptions in neurodevelopment throughout life \cite{De_Kieviet2012-bv}. In our study, we explore differences in WM changes between the full-term and preterm neonates to study the influence of birth status on brain development. In this analysis, we use data from the 134 preterm (age at birth $<32$ weeks PMA) and 307 full-term (age at birth $>= 32$ weeks PMA) neonates in the dHCP dataset. Then, within each group, for each tract, we perform a GLM analysis between the subject’s tract FA and gestational age to quantify the rate of WM change with age (i.e., the beta coefficient).

\section{Results}
\label{3}
\begin{figure}[htbp]
  \centering
  \begin{subfigure}[b]{1\textwidth}
    \centering
    \includegraphics[width=\textwidth]{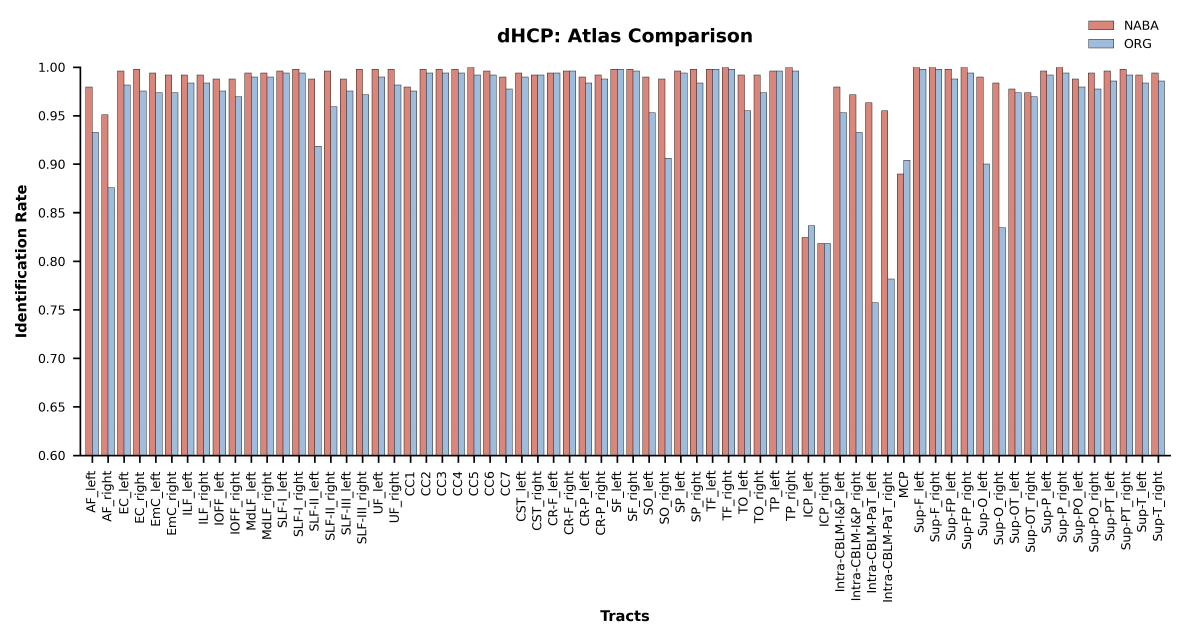}
    \caption{dHCP}
    \label{fig:subfig_a}
  \end{subfigure}
  \hfill
  \begin{subfigure}[b]{1\textwidth}
    \centering
    \includegraphics[width=\textwidth]{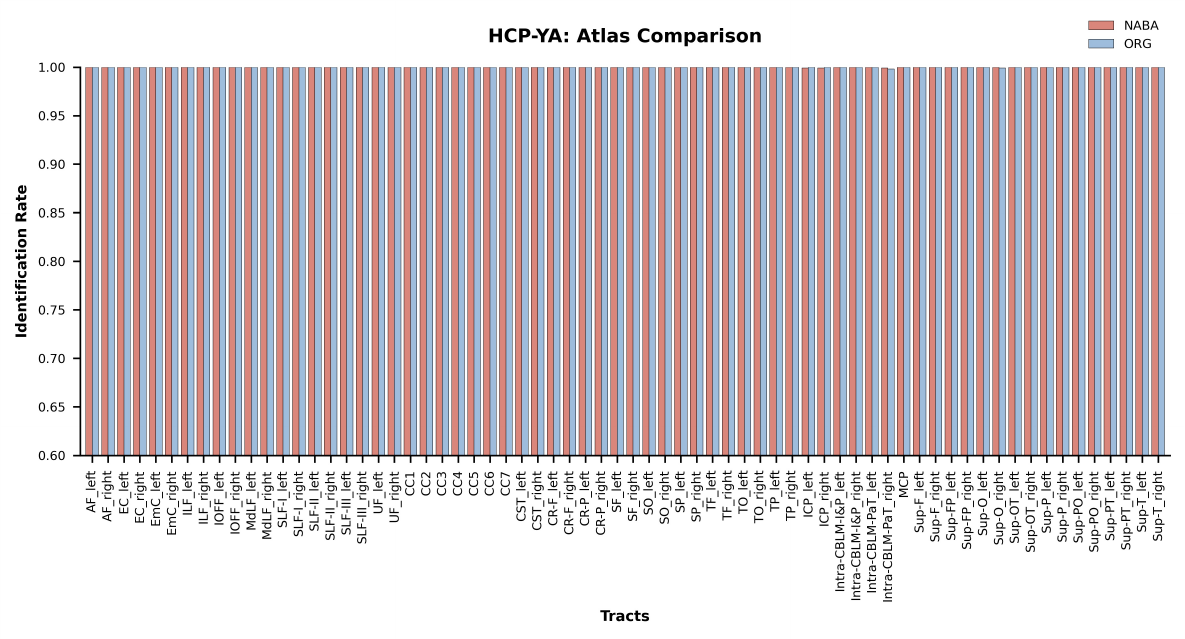}
    \caption{HCP-YA}
    \label{fig:subfig_b}
  \end{subfigure}
  \caption{IR comparison between NABA atlas and ORG atlas on dHCP and HCP-YA.}
  \label{fig:atlas comparison}
\end{figure}

{\sloppy
\subsection{Assessing the NABA Atlas for Concurrent WM Mapping between Neonate and Adult Brains}}
\label{3.1}
Figure \ref{fig:atlas comparison} presents a comparison of tract IRs of the dHCP and HCP-YA datasets between the ORG atlas and the NABA atlas. For the HCP-YA adult data, the average identification rates across all tracts are 99.996\% with the ORG atlas and the same with the NABA atlas. In contrast, for the dHCP neonate data, the NABA atlas achieves an average IR of 98.553\%, outperforming the ORG atlas, which achieves 96.522\%. A t-test on the IRs across all tracts shows a statistically significant improvement in favor of NABA ($p<0.0001$). These findings demonstrate that the NABA atlas significantly enhances tract detection performance in neonate data while maintaining comparable results to the ORG atlas for adult data.

\begin{figure}[htbp]
  \centering
  \includegraphics[width=1\textwidth]{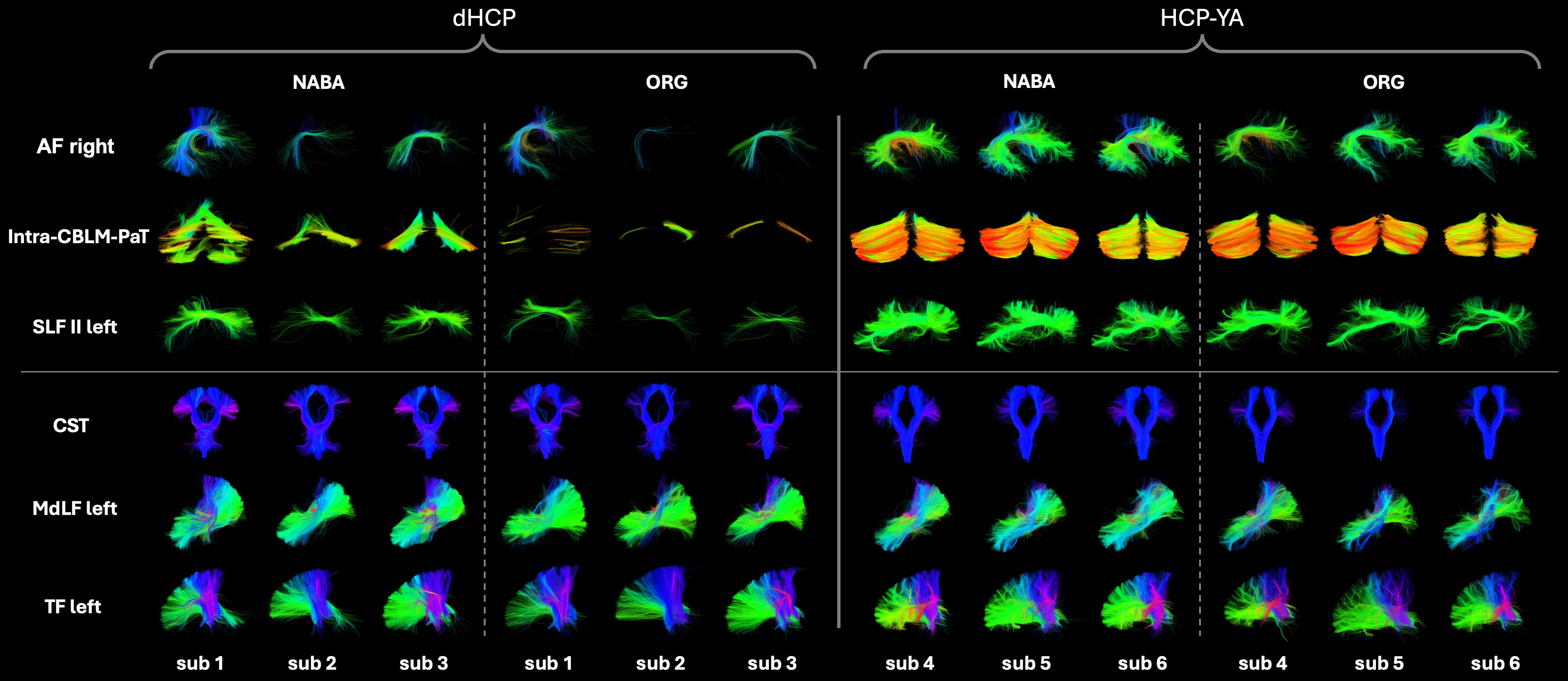}
  \caption{Visualization of the tract parcellation results with the NABA atlas and ORG atlas. Six example tracts are visualized. For each tract, the parcellation results in three dHCP subjects (left panel) and three HCP-YA subjects (right panel) are visualized.}
  \label{tract visualization}
\end{figure}

Figure \ref{tract visualization} visually compares the tracts parcellated in example dHCP and HCP-YA subjects using the ORG and NABA atlases. The left panel of Figure \ref{tract visualization} highlights example tracts from dHCP subjects, showcasing those with the largest IR differences (i.e., AF\_right, Intra-CBLM-PaT, SLF-II\_left) and those with the smallest differences (i.e., CST, MdLF\_left, TF\_left). For tracts with the largest IR differences, the NABA atlas identifies more streamlines and yields more anatomically plausible structures, while tracts with minimal differences exhibit consistent patterns across both atlases. The right panel of Figure \ref{tract visualization} shows the above tracts in the HCP-YA data. We can observe that both atlases produce visually consistent and comparable tracts, demonstrating their robustness in adult data.

\subsection{Exploring Developmental Changes of Neonatal Brains with Respect to the Adult Brains}
\label{3.2}
\begin{figure}[htbp]
  \centering
  \includegraphics[width=1\textwidth]{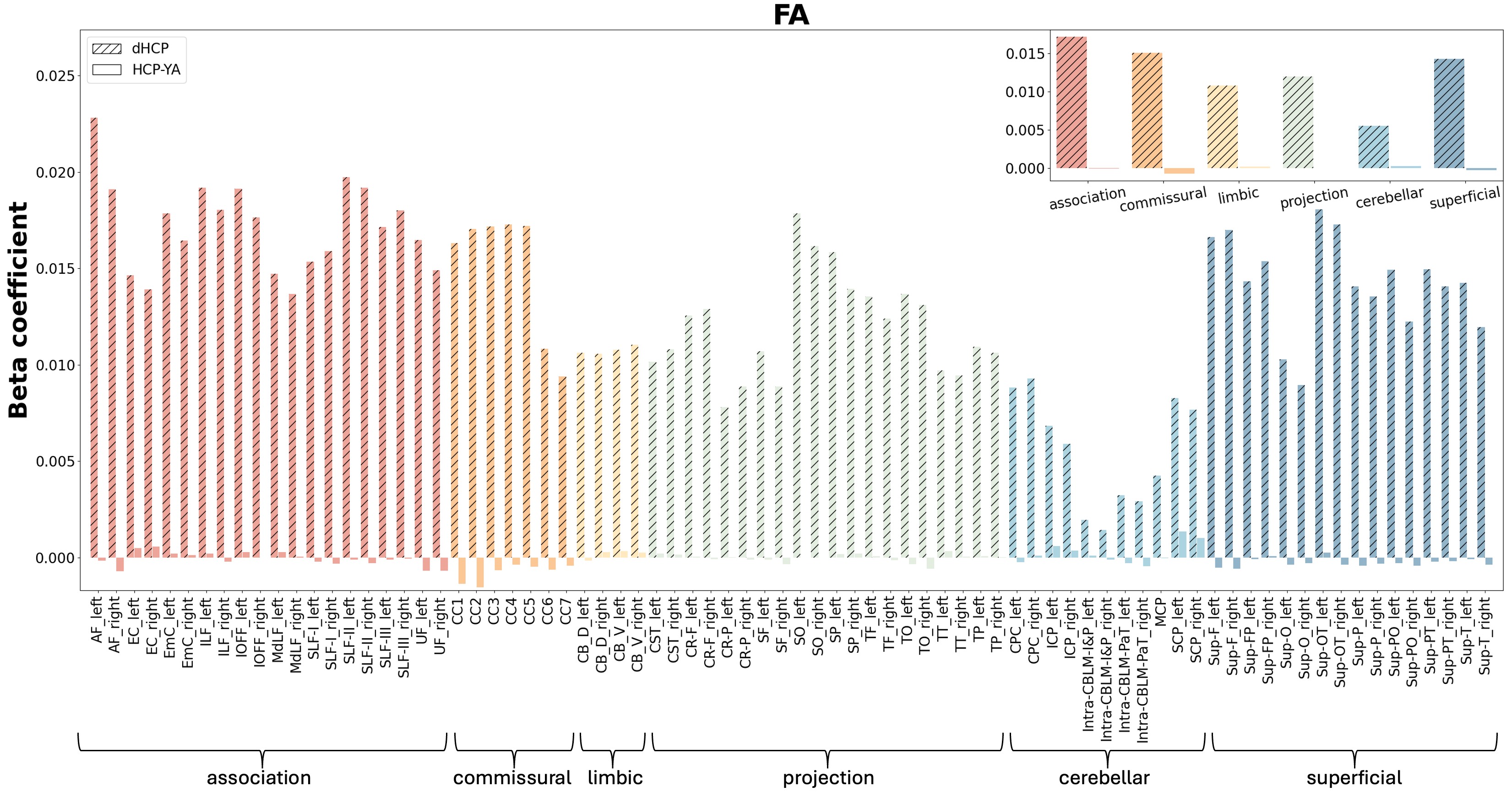}
  \caption{Rate of change with age (beta coefficient) of FA in each of the dHCP and HCP-YA populations. The two groups are distinguished using different marker styles for the bars. Tracts of different categories are color-coded. The upper right corner displays the rate of change at the category level for both populations.}
  \label{dHCP HCP beta}
\end{figure}

\begin{figure}[htbp]
  \centering
  \includegraphics[width=0.9\textwidth]{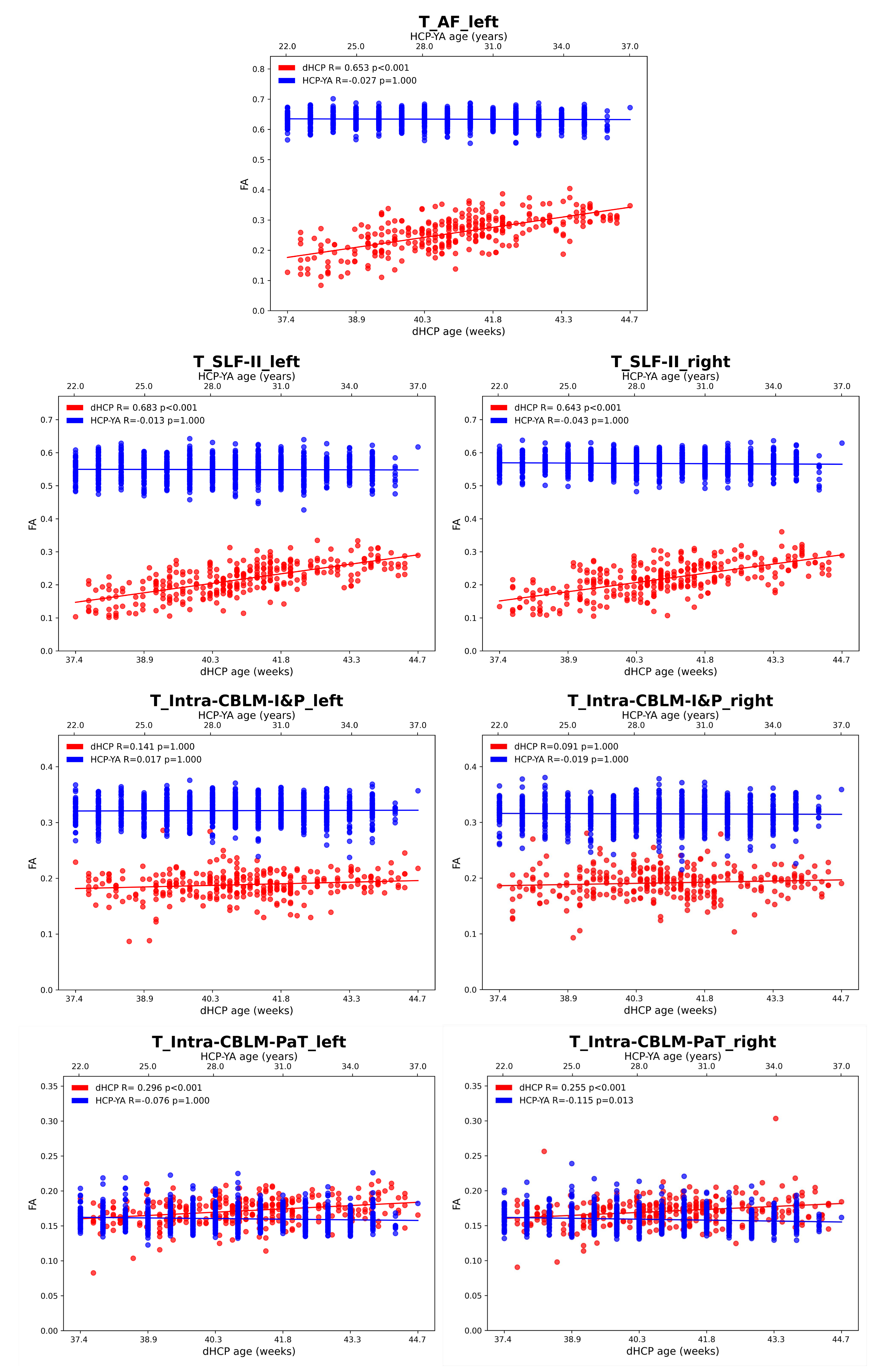}
  \caption{Changes in FA with increasing age in the AF, SLF II, and Intra-CBLM tracts.}
  \label{scatter}
\end{figure}

Figure \ref{dHCP HCP beta} gives the rate of FA change with age (i.e., the beta coefficient) for each fiber tract and each of the dHCP and HCP-YA datasets. In dHCP, all tracts show a significant FA increase with age ($p < 0.05$) except for the left and right Intra-CBLM-l\&P tracts. In contrast, within the HCP-YA age window (22–37 years), none of the tracts exhibit a significant increase with age. Tract category analysis reveals that, in the dHCP, association tracts generally exhibit the most rapid development, while cerebellar tracts show the slowest growth. Interestingly, the left AF and left SLF II tracts exhibit the highest beta coefficients in the dHCP data. In contrast, the intra-cerebellar tracts (I\&P and PaT) display the lowest beta coefficients. Figure \ref{scatter} shows the changes in FA with increasing age in these tracts.

\subsection{Investigating Sex Differences in Early WM Development}
\label{3.3}
\begin{figure}[htbp]
  \centering
  \includegraphics[width=0.9\textwidth]{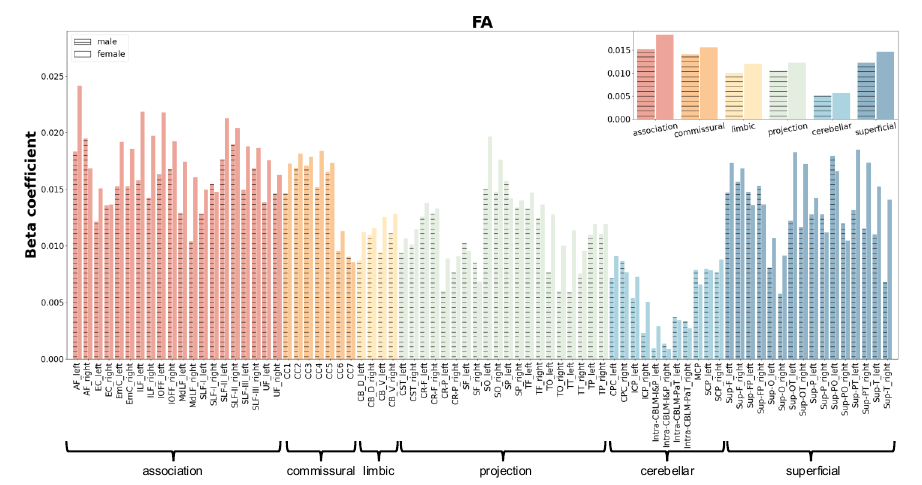}
  \caption{Sex differences in the rate of FA change with age (beta coefficient) of neonates in dHCP. The two groups (male and female) were balanced for birth age, and the GLM additionally controlled for birth weight and head circumference at scan. The two groups are distinguished using different marker styles for the bars. Tracts of different categories are color-coded. The upper right corner displays the rate of change at the category level for both populations.}
  \label{male female beta}
\end{figure}

Figure \ref{male female beta} presents the results of the sex difference comparison in the rate of FA change with age in dHCP. Overall, female neonates exhibit more rapid development across all tract categories. Across all individual tracts, there are a total of 61 tracts where female neonates exhibit higher beta coefficients and a total of 17 tracts where male neonates exhibit higher beta coefficients. These findings generally suggest that females experience faster WM development during the early stages of life after birth.

\subsection{Probing Developmental Differences due to Preterm Birth}
\label{3.4}
\begin{figure}[htbp]
  \centering
  \includegraphics[width=0.9\textwidth]{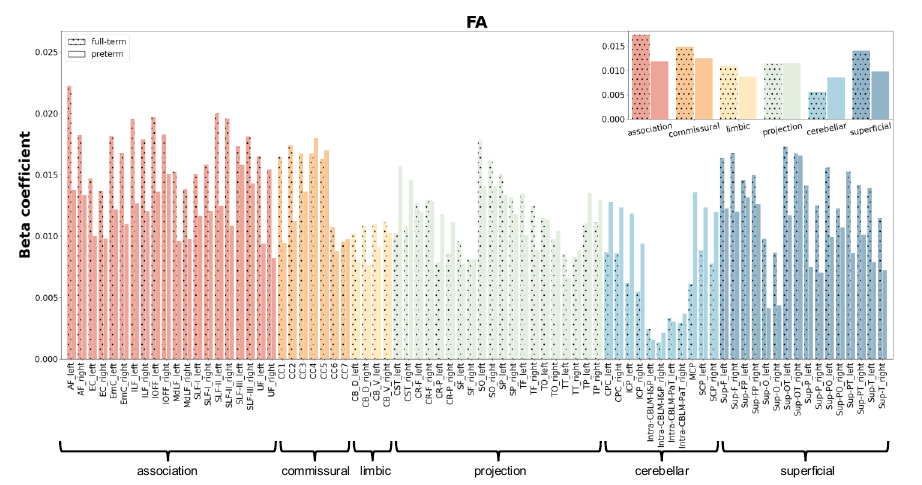}
  \caption{Rate of change with age (beta coefficient) of FA in full-term and preterm neonates in dHCP. The two groups were sex-balanced, and the GLM controlled for birth weight and head circumference at scan. The two groups are distinguished using different marker styles for the bars. Tracts of different categories are color-coded. The upper right corner displays the rate of change at the category level for both populations.}
  \label{fullterm preterm beta}
\end{figure}

Figure \ref{fullterm preterm beta} gives the results of the full-term vs preterm comparison in the rate of FA change with age in dHCP. The full-term subjects have significantly higher beta coefficients of FA than preterm subjects in association, commissural, limbic, and superficial tracts, with a similar beta coefficient in the projection tracts and a lower beta coefficient in the cerebellar tracts. Overall, in full-term subjects, we identify 58 tracts with higher beta coefficients, whereas 20 have lower beta coefficients. Among these 20 tracts, the ones showing the lowest beta coefficients are the CST, CR-P, and cerebellar tracts - all known to be motor-related pathways.

\section{Discussion}
This study proposes NABA, a novel WM tractography atlas constructed from a large-scale dMRI spanning both neonates and adults. By aligning the developing and adult{mature} brains into a common space with anatomical labeling, the NABA atlas enables concurrent mapping and direct comparison of WM across different developmental stages of human brain development. Through comprehensive experiments, NABA demonstrates superior performance in identifying WM tracts in neonates. We utilized it to examine the developmental trends of WM in neonates with respect to young adults. Furthermore, we investigate sex differences in development and the impact of preterm birth in neonates.

A key innovation of NABA is that it is a unified and population-averaged atlas that spans both neonates (scanned at  29.29-45.14 weeks PMA) and young adults (scanned at 22–37 years). It provides a common template that enables meaningful and anatomically consistent comparisons of WM tracts between neonates and young adults. There have been existing atlases created based on specific populations. But they often struggle to generalize across different age groups. NABA addresses this challenge through the following key points. First, an equal number of representative subjects are selected from both populations to participate in atlas creation with equivalent weighting. Tractography from both populations is registered into a common space. Second, a data-driven approach of tractography-based spectral clustering is performed to group morphologically similar fibers. These points ensure anatomically meaningful clustering and remain consistent across populations, enabling robust identification of tracts while preserving age-specific anatomical variations. Unlike previous neonatal studies that excluded the cerebellum due to substantial cross-age registration challenges, NABA supports whole-brain fiber analysis, enhancing anatomical coverage. Subsequent experiments further assessed NABA’s advantages, demonstrating its high performance in fiber tract identification across both populations.

In our first analysis, we evaluate NABA for concurrent WM mapping in both neonates and young adults. Compared to the ORG atlas that was created using only the adult data, our NABA atlas achieves highly improved tract detection performance in neonates. The quantitative results show significantly enhanced tract identification rates in tracts such as AF, SLF II, Intra-CBLM-PaT tract, and Sup-O. For example, NABA enables more complete and spatially coherent AF and SLF II tracts that are known to be associated with language. Visualization results demonstrate that NABA reconstructs tract geometries with greater anatomical plausibility compared to the ORG atlas, which yields sparser and fragmented streamlines. It should be noted that identifying fibers in neonates remains challenging due to the immature microstructural properties of neonatal WM. Specifically, low myelination and reduced anisotropy in neonates lead to insufficient directional diffusion signals for tractography \cite{dubois2021mri}, resulting in inherently lower IR compared to adults, even when using an atlas constructed for this age group. On the other hand, in the HCP-YA data, we show that NABA can maintain comparable results in young adults with the ORG atlas in terms of quantitative tract IR and visualization of tract shape. We also calculate the mean microstructure FA values across all tracts detected in each atlas and find highly similar values (0.495 vs 0.498). Overall, these results indicate that NABA offers better adaptability to age-specific anatomical variability, especially in tracts that undergo considerable structural changes across development.

In our second analysis, we examine the rate of change of FA across neonates and young adults. We observe widespread and significant FA increases with age among neonates. Association tracts, such as the left AF and the left SLF II, known to be language-related \cite{Catani2008-hz,Ivanova2016-iq,Sander2022-cr}, exhibit the most rapid developmental changes. This aligns with previous findings \cite{Geng2012-lv,Sket2019-sx}, where AF is relatively underdeveloped at birth but develops rapidly during the first year of life compared to other tracts. For the left SLF II, Liang et al. find that SLF II exhibits the lowest developmental level among the three subdivisions of SLF in neonates \cite{Liang2022-el}. Our analysis not only identifies this but also shows that SLF II has a more rapid development trend. The intra-cerebellar tracts (I\&P and PaT), known to be majorly associated with motor functions \cite{Beez2021-gg,Benagiano2018-jz,De_Benedictis2022-kc}, display the lowest beta coefficients. However, the FA of them is closer to the young adult level. This may indicate earlier development of the cerebellum, which aligns with the findings from Deoni et al. that the cerebellum is one of the earliest regions to undergo myelination \cite{Deoni2012-yj}. As tracts associated with motor and cognitive function, their continued growth may reflect the relatively underdeveloped state of motor and cognitive abilities at birth \cite{Manto2012-yh,Schmahmann2010-td}. Across the lifespan, Lebel et al report that FA increases during childhood and adolescence, reaches a peak between 20 and 42 years of age, and then decreases \cite{Lebel2012-rc} non-linear changes of FA as age increases during adulthood. To investigate this, we also fit the FA with birth age in years to a Poisson curve to capture potential nonlinear developmental patterns following the analysis in  \cite{Lebel2012-rc}, but we do not find any statistical significance. In our study, our results show no significant growth of FA in any tracts under study, possibly due to the short age range in the HCP-YA cohort.

We further examine sex-related differences in neonatal WM development. Female neonates exhibit generally faster FA increases across most tract categories compared to age-matched male neonates, except in cerebellar tracts. Across individual tracts, females show higher beta coefficients in most tracts than males. I the existing studies, Deoni et al. report that between 3 months and 5 years of age, females show faster development of myelin water fraction (MWF) than males in the genu of the corpus callosum, left frontal and temporal white matter, and right optic radiations \cite{Deoni2012-yj}. Geng et al. report that males exhibit lower FA than females in the right sensory tract among children aged 0 to 2 years \cite{Geng2012-lv}. Two review articles also conclude that there are sex differences in brain development that persist to a certain age \cite{Buyanova2021-pa,Saker2024-tn}. In our analysis, the overall faster FA growth rate observed in females may indicate sex-specific trends in brain development. These results may have implications for understanding sex-based differences in neurodevelopment and call for further longitudinal investigation.

Our final analysis investigates the effects of preterm birth on WM development{maturation}. We find the FA of tracts in preterm infants is overall lower than in full-term infants. It suggests that preterm birth may lead to delayed development. Previous research findings similarly support this point. For example, Lee et al. find preterm infants have lower WM FA values than full-term infants \cite{Lee2013-km}. Thompson et al. find that FA was lower in the corpus callosum of very preterm infants compared to full-term infants \cite{Thompson2011-rx}. Two TBSS studies find that preterm infants show lower FA in the genu and splenium of the corpus callosum, frontal white matter, centrum semiovale, external capsule, and corona radiata \cite{Anjari2007-te,Rose2008-cn}. Compared to full-term neonates, our analysis also finds that preterm infants exhibit lower FA growth rates in the majority of tracts, including association, commissural, limbic, and superficial tracts. This suggests delayed or disrupted development in these regions. Notably, our results show that certain motor-related tracts, including CST, CR-P, and intra-cerebellar tracts, show relatively higher FA growth rates in preterm infants. To complement the FA analysis, we also investigate the mean diffusivity (MD) of the two groups, as shown in Figures S4 and S5. Consistent with the FA results, the overall trend of decreasing MD points to accelerated white matter development in preterm infants. The smaller effect sizes (beta changes) for MD compared to FA, however, indicate that FA is likely a more sensitive biomarker for capturing developmental differences between preterm and full-term cohorts. Several studies in preterm populations have suggested similar tendencies. Giménez et al. found higher FA at term-equivalent age in preterm infants, possibly reflecting accelerated maturation in sensorimotor pathways \cite{gimenez2008accelerated}. Kimpton et al. and Berman et al. also reported rapid maturation of the corticospinal and related sensorimotor tracts in this population \cite{kimpton2021diffusion,berman2005quantitative}. In addition, diffusion measures in the cerebellar peduncles have been associated with early motor performance, indicating close coupling between cerebellar and motor systems during development \cite{shany2017diffusion}.
Overall, these findings suggest that although global WM maturation in preterm infants is delayed, motor-related tracts may develop relatively faster, possibly influenced by early postnatal sensorimotor activity. Further longitudinal work is needed to clarify whether this represents true acceleration or transient adaptation. The NABA atlas enables detailed tract-wise investigation of these differences. It may aid future studies in understanding the developmental mechanisms underlying preterm-related alterations in brain connectivity.

While NABA demonstrates robust performance in identifying WM tracts across age groups, several limitations should be noted. First, the current atlas is constructed using diffusion MRI data from only two age groups: neonates from the dHCP and young adults from the HCP-YA. A future direction will be to extend the atlas to include high-quality lifespan datasets, covering a broader range of age groups, and thus enabling the investigation of WM development across the full lifespan. Second, the tract parcellation and anatomical labeling rely on unsupervised spectral clustering that can be time-consuming. Future work could include the application of deep learning for fast atlas creation using a larger scale of data. Third, this study primarily uses FA for group-wise analysis. While our supplementary MD findings show a similar trend, differences exist due to their distinct biological underpinnings. Future work could systematically compare multiple diffusion metrics to elucidate their commonalities and differences.

\section{Conclusion}
This study proposes NABA, a novel cross-population brain WM atlas that enables concurrent mapping and direct anatomical comparison of WM between neonates and adults. NABA offers a standardized framework and toolset for studying the WM across different populations and developmental stages, which overcomes limitations of traditional population-specific atlases and significantly improves tract identification, particularly in neonatal data. By using NABA, our analyses reveal biologically meaningful developmental trends. We show that tract-specific differences in neonatal WM development compared to young adults, and that both sex and preterm birth influence WM FA development. These findings not only affirm the sensitivity of NABA for capturing subtle developmental differences but also contribute novel insights into early brain development patterns. Importantly, NABA offers an effective framework for investigating white matter across different developmental stages, even in cases of atypical development such as developmental disorders. Although this study primarily focuses on neonates and young adults, extending NABA to cover the full lifespan may help uncover the continuous trajectory of white matter changes throughout life. 

\section{Acknowledgments}
This work is in part supported by the National Natural Science Foundation of China [grant number 62371107] and the National Key R\&D Program of China [grant number 2023YFE0118600].

\bibliographystyle{elsarticle-num}
\bibliography{references}

@ARTICLE{Geng2012-lv,
  title     = "Quantitative tract-based white matter development from birth to
               age {2years}",
  author    = "Geng, Xiujuan and Gouttard, Sylvain and Sharma, Anuja and Gu,
               Hongbin and Styner, Martin and Lin, Weili and Gerig, Guido and
               Gilmore, John H",
  journal   = "Neuroimage",
  publisher = "Elsevier BV",
  volume    =  61,
  number    =  3,
  pages     = "542--557",
  month     =  jul,
  year      =  2012,
  language  = "en"
}

@ARTICLE{Xue2023-km,
  title    = "Superficial white matter analysis: An efficient point-cloud-based
              deep learning framework with supervised contrastive learning for
              consistent tractography parcellation across populations and {dMRI}
              acquisitions",
  author   = "Xue, Tengfei and Zhang, Fan and Zhang, Chaoyi and Chen, Yuqian and
              Song, Yang and Golby, Alexandra J and Makris, Nikos and Rathi,
              Yogesh and Cai, Weidong and O'Donnell, Lauren J",
  journal  = "Med. Image Anal.",
  volume   =  85,
  pages    =  102759,
  month    =  apr,
  year     =  2023,
  language = "en"
}

@ARTICLE{Roman2017-hd,
  title    = "Clustering of Whole-Brain White Matter Short Association Bundles
              Using {HARDI} Data",
  author   = "Román, Claudio and Guevara, Miguel and Valenzuela, Ronald and
              Figueroa, Miguel and Houenou, Josselin and Duclap, Delphine and
              Poupon, Cyril and Mangin, Jean-François and Guevara, Pamela",
  journal  = "Front. Neuroinform.",
  volume   =  11,
  pages    =  73,
  month    =  dec,
  year     =  2017,
  language = "en"
}

@ARTICLE{Verschuur2024-gq,
  title     = "Methodological considerations on diffusion {MRI} tractography in
               infants aged 0-2 years: a scoping review",
  author    = "Verschuur, Anouk S and King, Regan and Tax, Chantal M W and
               Boomsma, Martijn F and van Wezel-Meijler, Gerda and Leemans,
               Alexander and Leijser, Lara M",
  journal   = "Pediatr. Res.",
  publisher = "Springer Science and Business Media LLC",
  pages     = "1--18",
  month     =  aug,
  year      =  2024,
  language  = "en"
}

@ARTICLE{Dubois2014-my,
  title     = "The early development of brain white matter: a review of imaging
               studies in fetuses, newborns and infants",
  author    = "Dubois, J and Dehaene-Lambertz, G and Kulikova, S and Poupon, C
               and Hüppi, P S and Hertz-Pannier, L",
  journal   = "Neuroscience",
  publisher = "Elsevier BV",
  volume    =  276,
  pages     = "48--71",
  month     =  sep,
  year      =  2014,
  language  = "en"
}

@ARTICLE{Cabral2022-cz,
  title    = "Anatomical correlates of category-selective visual regions have
              distinctive signatures of connectivity in neonates",
  author   = "Cabral, Laura and Zubiaurre-Elorza, Leire and Wild, Conor J and
              Linke, Annika and Cusack, Rhodri",
  journal  = "Dev. Cogn. Neurosci.",
  volume   =  58,
  pages    =  101179,
  month    =  dec,
  year     =  2022,
  language = "en"
}

@ARTICLE{ODonnell2007-zx,
  title    = "Automatic tractography segmentation using a high-dimensional white
              matter atlas",
  author   = "O'Donnell, Lauren J and Westin, Carl-Fredrik",
  journal  = "IEEE Trans. Med. Imaging",
  volume   =  26,
  number   =  11,
  pages    = "1562--1575",
  month    =  nov,
  year     =  2007,
  language = "en"
}

@ARTICLE{Saker2024-tn,
  title     = "Insight into brain sex differences of typically developed infants
               and brain pathologies: A systematic review",
  author    = "Saker, Zahraa and Rizk, Mahdi and Merie, Diana and Nabha, Rami H
               and Pariseau, Nicole J and Nabha, Sanaa M and Makki, Malek I",
  journal   = "Eur. J. Neurosci.",
  publisher = "Wiley",
  volume    =  60,
  number    =  1,
  pages     = "3491--3504",
  month     =  jul,
  year      =  2024,
  language  = "en"
}

@ARTICLE{Gaillard2001-pq,
  title     = "Developmental aspects of pediatric {fMRI}: considerations for
               image acquisition, analysis, and interpretation",
  author    = "Gaillard, W D and Grandin, C B and Xu, B",
  journal   = "Neuroimage",
  publisher = "Elsevier BV",
  volume    =  13,
  number    =  2,
  pages     = "239--249",
  month     =  feb,
  year      =  2001,
  language  = "en"
}

@ARTICLE{Ouyang2019-vj,
  title    = "Delineation of early brain development from fetuses to infants
              with diffusion {MRI} and beyond",
  author   = "Ouyang, Minhui and Dubois, Jessica and Yu, Qinlin and Mukherjee,
              Pratik and Huang, Hao",
  journal  = "Neuroimage",
  volume   =  185,
  pages    = "836--850",
  month    =  jan,
  year     =  2019,
  language = "en"
}

@ARTICLE{Wilkinson2017-yw,
  title    = "Detection and Growth Pattern of Arcuate Fasciculus from Newborn to
              Adult",
  author   = "Wilkinson, Molly and Lim, Ashley R and Cohen, Andrew H and
              Galaburda, Albert M and Takahashi, Emi",
  journal  = "Front. Neurosci.",
  volume   =  11,
  pages    =  389,
  month    =  jul,
  year     =  2017,
  language = "en"
}

@ARTICLE{Yeh2018-cx,
  title    = "Population-averaged atlas of the macroscale human structural
              connectome and its network topology",
  author   = "Yeh, Fang-Cheng and Panesar, Sandip and Fernandes, David and
              Meola, Antonio and Yoshino, Masanori and Fernandez-Miranda, Juan C
              and Vettel, Jean M and Verstynen, Timothy",
  journal  = "Neuroimage",
  volume   =  178,
  pages    = "57--68",
  month    =  sep,
  year     =  2018,
  language = "en"
}

@ARTICLE{Huang2006-yd,
  title     = "White and gray matter development in human fetal, newborn and
               pediatric brains",
  author    = "Huang, Hao and Zhang, Jiangyang and Wakana, Setsu and Zhang,
               Weihong and Ren, Tianbo and Richards, Linda J and Yarowsky, Paul
               and Donohue, Pamela and Graham, Ernest and van Zijl, Peter C M
               and Mori, Susumu",
  journal   = "Neuroimage",
  publisher = "Elsevier BV",
  volume    =  33,
  number    =  1,
  pages     = "27--38",
  month     =  oct,
  year      =  2006,
  language  = "en"
}

@ARTICLE{Tak2016-uw,
  title     = "Developmental process of the arcuate fasciculus from infancy to
               adolescence: a diffusion tensor imaging study",
  author    = "Tak, Hyeong Jun and Kim, Jin Hyun and Son, Su Min",
  journal   = "Neural Regen. Res.",
  publisher = "Medknow",
  volume    =  11,
  number    =  6,
  pages     = "937--943",
  month     =  jun,
  year      =  2016,
  language  = "en"
}

@ARTICLE{Catani2008-hz,
  title     = "The arcuate fasciculus and the disconnection theme in language
               and aphasia: history and current state",
  author    = "Catani, Marco and Mesulam, Marsel",
  journal   = "Cortex",
  publisher = "Elsevier BV",
  volume    =  44,
  number    =  8,
  pages     = "953--961",
  month     =  sep,
  year      =  2008,
  language  = "en"
}

@ARTICLE{Ivanova2016-iq,
  title    = "Diffusion-tensor imaging of major white matter tracts and their
              role in language processing in aphasia",
  author   = "Ivanova, Maria V and Isaev, Dmitry Yu and Dragoy, Olga V and
              Akinina, Yulia S and Petrushevskiy, Alexey G and Fedina, Oksana N
              and Shklovsky, Victor M and Dronkers, Nina F",
  journal  = "Cortex",
  volume   =  85,
  pages    = "165--181",
  month    =  dec,
  year     =  2016,
  language = "en"
}

@ARTICLE{Lebel2012-rc,
  title     = "Diffusion tensor imaging of white matter tract evolution over the
               lifespan",
  author    = "Lebel, C and Gee, M and Camicioli, R and Wieler, M and Martin, W
               and Beaulieu, C",
  journal   = "Neuroimage",
  publisher = "Elsevier BV",
  volume    =  60,
  number    =  1,
  pages     = "340--352",
  month     =  mar,
  year      =  2012,
  language  = "en"
}

@ARTICLE{Basser1994-mh,
  title     = "{MR} diffusion tensor spectroscopy and imaging",
  author    = "Basser, P J and Mattiello, J and LeBihan, D",
  journal   = "Biophys. J.",
  publisher = "ui.adsabs.harvard.edu",
  volume    =  66,
  number    =  1,
  pages     = "259--267",
  month     =  jan,
  year      =  1994,
  language  = "en"
}

@ARTICLE{Liang2022-el,
  title    = "A comparative study of the superior longitudinal fasciculus
              subdivisions between neonates and young adults",
  author   = "Liang, Wenjia and Yu, Qiaowen and Wang, Wenjun and Dhollander,
              Thijs and Suluba, Emmanuel and Li, Zhuoran and Xu, Feifei and Hu,
              Yang and Tang, Yuchun and Liu, Shuwei",
  journal  = "Brain Struct. Funct.",
  volume   =  227,
  number   =  8,
  pages    = "2713--2730",
  month    =  nov,
  year     =  2022,
  language = "en"
}

@ARTICLE{ODonnell2012-kp,
  title    = "Unbiased groupwise registration of white matter tractography",
  author   = "O'Donnell, Lauren J and Wells, 3rd, William M and Golby, Alexandra
              J and Westin, Carl-Fredrik",
  journal  = "Med. Image Comput. Comput. Assist. Interv.",
  volume   =  15,
  number   = "Pt 3",
  pages    = "123--130",
  year     =  2012,
  language = "en"
}

@ARTICLE{Truzzi2023-np,
  title    = "The development of intrinsic timescales: A comparison between the
              neonate and adult brain",
  author   = "Truzzi, Anna and Cusack, Rhodri",
  journal  = "Neuroimage",
  volume   =  275,
  pages    =  120155,
  month    =  jul,
  year     =  2023,
  language = "en"
}

@ARTICLE{Asato2010-xs,
  title     = "White matter development in adolescence: a {DTI} study",
  author    = "Asato, M R and Terwilliger, R and Woo, J and Luna, B",
  journal   = "Cereb. Cortex",
  publisher = "Oxford University Press (OUP)",
  volume    =  20,
  number    =  9,
  pages     = "2122--2131",
  month     =  sep,
  year      =  2010,
  language  = "en"
}

@ARTICLE{Makropoulos2016-oc,
  title    = "Regional growth and atlasing of the developing human brain",
  author   = "Makropoulos, Antonios and Aljabar, Paul and Wright, Robert and
              Hüning, Britta and Merchant, Nazakat and Arichi, Tomoki and Tusor,
              Nora and Hajnal, Joseph V and Edwards, A David and Counsell,
              Serena J and Rueckert, Daniel",
  journal  = "Neuroimage",
  volume   =  125,
  pages    = "456--478",
  month    =  jan,
  year     =  2016,
  language = "en"
}

@ARTICLE{Anjari2007-te,
  title   = "Diffusion tensor imaging with tract-based spatial statistics
             reveals local white matter abnormalities in preterm infants",
  author  = "Anjari, M and Srinivasan, L and Allsop, J M and Hajnal, J V and
             Rutherford, M A and Edwards, A D and Counsell, S J",
  journal = "Neuroimage",
  volume  =  35,
  number  =  3,
  pages   = "1021--1027",
  year    =  2007
}

@ARTICLE{Basser2000-gg,
  title    = "In vivo fiber tractography using {DT}-{MRI} data",
  author   = "Basser, P J and Pajevic, S and Pierpaoli, C and Duda, J and
              Aldroubi, A",
  journal  = "Magn. Reson. Med.",
  volume   =  44,
  number   =  4,
  pages    = "625--632",
  month    =  oct,
  year     =  2000,
  language = "en"
}

@ARTICLE{Li2024-kr,
  title     = "A diffusion {MRI} tractography atlas for concurrent white matter
               mapping across Eastern and Western populations",
  author    = "Li, Yijie and Zhang, Wei and Wu, Ye and Yin, Li and Zhu, Ce and
               Chen, Yuqian and Cetin-Karayumak, Suheyla and Cho, Kang Ik K and
               Zekelman, Leo R and Rushmore, Jarrett and Rathi, Yogesh and
               Makris, Nikos and O'Donnell, Lauren J and Zhang, Fan",
  journal   = "Scientific Data",
  publisher = "Nature Publishing Group",
  volume    =  11,
  number    =  1,
  pages     = "1--14",
  month     =  jul,
  year      =  2024,
  language  = "en"
}

@ARTICLE{Beez2021-gg,
  title     = "Functional tracts of the cerebellum-essentials for the
               neurosurgeon",
  author    = "Beez, Thomas and Munoz-Bendix, Christopher and Steiger,
               Hans-Jakob and Hänggi, Daniel",
  journal   = "Neurosurg. Rev.",
  publisher = "Springer Science and Business Media LLC",
  volume    =  44,
  number    =  1,
  pages     = "273--278",
  month     =  feb,
  year      =  2021,
  language  = "en"
}

@ARTICLE{Zhai2003-ma,
  title    = "Comparisons of regional white matter diffusion in healthy neonates
              and adults performed with a 3.0-{T} head-only {MR} imaging unit",
  author   = "Zhai, Guihua and Lin, Weili and Wilber, Kathy P and Gerig, Guido
              and Gilmore, John H",
  journal  = "Radiology",
  volume   =  229,
  number   =  3,
  pages    = "673--681",
  month    =  dec,
  year     =  2003,
  language = "en"
}

@ARTICLE{ODonnell2017-fy,
  title    = "Automated white matter fiber tract identification in patients with
              brain tumors",
  author   = "O'Donnell, Lauren J and Suter, Yannick and Rigolo, Laura and
              Kahali, Pegah and Zhang, Fan and Norton, Isaiah and Albi, Angela
              and Olubiyi, Olutayo and Meola, Antonio and Essayed, Walid I and
              Unadkat, Prashin and Ciris, Pelin Aksit and Wells, 3rd, William M
              and Rathi, Yogesh and Westin, Carl-Fredrik and Golby, Alexandra J",
  journal  = "Neuroimage Clin",
  volume   =  13,
  pages    = "138--153",
  year     =  2017,
  language = "en"
}

@ARTICLE{Guevara2012-id,
  title    = "Automatic fiber bundle segmentation in massive tractography
              datasets using a multi-subject bundle atlas",
  author   = "Guevara, P and Duclap, D and Poupon, C and Marrakchi-Kacem, L and
              Fillard, P and Le Bihan, D and Leboyer, M and Houenou, J and
              Mangin, J-F",
  journal  = "Neuroimage",
  volume   =  61,
  number   =  4,
  pages    = "1083--1099",
  month    =  jul,
  year     =  2012,
  language = "en"
}

@INPROCEEDINGS{Moberts2005-yk,
  title     = "Evaluation of fiber clustering methods for diffusion tensor
               imaging",
  author    = "Moberts, B and Vilanova, A and van Wijk, J J",
  booktitle = "VIS 05. IEEE Visualization, 2005.",
  pages     = "65--72",
  month     =  oct,
  year      =  2005
}

@ARTICLE{Laird2009-hk,
  title    = "{ALE} Meta-Analysis Workflows Via the Brainmap Database: Progress
              Towards A Probabilistic Functional Brain Atlas",
  author   = "Laird, Angela R and Eickhoff, Simon B and Kurth, Florian and Fox,
              Peter M and Uecker, Angela M and Turner, Jessica A and Robinson,
              Jennifer L and Lancaster, Jack L and Fox, Peter T",
  journal  = "Front. Neuroinform.",
  volume   =  3,
  pages    =  23,
  month    =  jul,
  year     =  2009,
  language = "en"
}

@ARTICLE{Lebel2018-ad,
  title    = "The development of brain white matter microstructure",
  author   = "Lebel, Catherine and Deoni, Sean",
  journal  = "Neuroimage",
  volume   =  182,
  pages    = "207--218",
  month    =  nov,
  year     =  2018,
  language = "en"
}

@ARTICLE{De_Benedictis2022-kc,
  title    = "Networking of the human cerebellum: From anatomo-functional
              development to neurosurgical implications",
  author   = "De Benedictis, Alessandro and Rossi-Espagnet, Maria Camilla and de
              Palma, Luca and Carai, Andrea and Marras, Carlo Efisio",
  journal  = "Front. Neurol.",
  volume   =  13,
  pages    =  806298,
  month    =  feb,
  year     =  2022,
  language = "en"
}

@ARTICLE{Zhang2020-nh,
  title    = "Creation of a novel trigeminal tractography atlas for automated
              trigeminal nerve identification",
  author   = "Zhang, Fan and Xie, Guoqiang and Leung, Laura and Mooney, Michael
              A and Epprecht, Lorenz and Norton, Isaiah and Rathi, Yogesh and
              Kikinis, Ron and Al-Mefty, Ossama and Makris, Nikos and Golby,
              Alexandra J and O'Donnell, Lauren J",
  journal  = "Neuroimage",
  volume   =  220,
  pages    =  117063,
  month    =  oct,
  year     =  2020,
  language = "en"
}

@ARTICLE{Rose2008-cn,
  title     = "Altered white matter diffusion anisotropy in normal and preterm
               infants at term-equivalent age",
  author    = "Rose, Stephen E and Hatzigeorgiou, Xanthy and Strudwick, Mark W
               and Durbridge, Gail and Davies, Peter S W and Colditz, Paul B",
  journal   = "Magn. Reson. Med.",
  publisher = "Wiley",
  volume    =  60,
  number    =  4,
  pages     = "761--767",
  month     =  oct,
  year      =  2008,
  language  = "en"
}

@ARTICLE{Farquharson2013-xc,
  title    = "White matter fiber tractography: why we need to move beyond {DTI}",
  author   = "Farquharson, Shawna and Tournier, J-Donald and Calamante, Fernando
              and Fabinyi, Gavin and Schneider-Kolsky, Michal and Jackson,
              Graeme D and Connelly, Alan",
  journal  = "J. Neurosurg.",
  volume   =  118,
  number   =  6,
  pages    = "1367--1377",
  month    =  jun,
  year     =  2013,
  language = "en"
}

@ARTICLE{Thompson2011-rx,
  title     = "Characterization of the corpus callosum in very preterm and
               full-term infants utilizing {MRI}",
  author    = "Thompson, Deanne K and Inder, Terrie E and Faggian, Nathan and
               Johnston, Leigh and Warfield, Simon K and Anderson, Peter J and
               Doyle, Lex W and Egan, Gary F",
  journal   = "Neuroimage",
  publisher = "Elsevier BV",
  volume    =  55,
  number    =  2,
  pages     = "479--490",
  month     =  mar,
  year      =  2011,
  language  = "en"
}

@ARTICLE{Makropoulos2018-vj,
  title    = "The developing human connectome project: A minimal processing
              pipeline for neonatal cortical surface reconstruction",
  author   = "Makropoulos, Antonios and Robinson, Emma C and Schuh, Andreas and
              Wright, Robert and Fitzgibbon, Sean and Bozek, Jelena and
              Counsell, Serena J and Steinweg, Johannes and Vecchiato, Katy and
              Passerat-Palmbach, Jonathan and Lenz, Gregor and Mortari, Filippo
              and Tenev, Tencho and Duff, Eugene P and Bastiani, Matteo and
              Cordero-Grande, Lucilio and Hughes, Emer and Tusor, Nora and
              Tournier, Jacques-Donald and Hutter, Jana and Price, Anthony N and
              Teixeira, Rui Pedro A G and Murgasova, Maria and Victor, Suresh
              and Kelly, Christopher and Rutherford, Mary A and Smith, Stephen M
              and Edwards, A David and Hajnal, Joseph V and Jenkinson, Mark and
              Rueckert, Daniel",
  journal  = "Neuroimage",
  volume   =  173,
  pages    = "88--112",
  month    =  jun,
  year     =  2018,
  language = "en"
}

@ARTICLE{Collins1994-vo,
  title    = "Automatic {3D} intersubject registration of {MR} volumetric data
              in standardized Talairach space",
  author   = "Collins, D L and Neelin, P and Peters, T M and Evans, A C",
  journal  = "J. Comput. Assist. Tomogr.",
  volume   =  18,
  number   =  2,
  pages    = "192--205",
  year     =  1994,
  language = "en"
}

@ARTICLE{Sander2022-cr,
  title     = "Frontoparietal anatomical connectivity predicts second language
               learning success",
  author    = "Sander, Kaija and Barbeau, Elise B and Chai, Xiaoqian and
               Kousaie, Shanna and Petrides, Michael and Baum, Shari and Klein,
               Denise",
  journal   = "Cereb. Cortex",
  publisher = "Oxford University Press (OUP)",
  volume    =  32,
  number    =  12,
  pages     = "2602--2610",
  month     =  jun,
  year      =  2022,
  language  = "en"
}

@ARTICLE{Zhang2022-he,
  title    = "Quantitative mapping of the brain's structural connectivity using
              diffusion {MRI} tractography: A review",
  author   = "Zhang, Fan and Daducci, Alessandro and He, Yong and Schiavi,
              Simona and Seguin, Caio and Smith, Robert E and Yeh, Chun-Hung and
              Zhao, Tengda and O'Donnell, Lauren J",
  journal  = "Neuroimage",
  volume   =  249,
  pages    =  118870,
  month    =  apr,
  year     =  2022,
  language = "en"
}

@ARTICLE{Zhang2018-lx,
  title    = "An anatomically curated fiber clustering white matter atlas for
              consistent white matter tract parcellation across the lifespan",
  author   = "Zhang, Fan and Wu, Ye and Norton, Isaiah and Rigolo, Laura and
              Rathi, Yogesh and Makris, Nikos and O'Donnell, Lauren J",
  journal  = "Neuroimage",
  volume   =  179,
  pages    = "429--447",
  month    =  oct,
  year     =  2018,
  language = "en"
}

@ARTICLE{Kazemi2007-lp,
  title     = "A neonatal atlas template for spatial normalization of
               whole-brain magnetic resonance images of newborns: preliminary
               results",
  author    = "Kazemi, Kamran and Moghaddam, Hamid Abrishami and Grebe, Reinhard
               and Gondry-Jouet, Catherine and Wallois, Fabrice",
  journal   = "Neuroimage",
  publisher = "Elsevier BV",
  volume    =  37,
  number    =  2,
  pages     = "463--473",
  month     =  aug,
  year      =  2007,
  language  = "en"
}

@ARTICLE{Buyanova2021-pa,
  title     = "Cerebral white matter myelination and relations to age, gender,
               and cognition: A selective review",
  author    = "Buyanova, Irina S and Arsalidou, Marie",
  journal   = "Front. Hum. Neurosci.",
  publisher = "Frontiers Media SA",
  volume    =  15,
  pages     =  662031,
  month     =  jul,
  year      =  2021,
  language  = "en"
}

@ARTICLE{Warrington2022-dp,
  title    = "Concurrent mapping of brain ontogeny and phylogeny within a common
              space: Standardized tractography and applications",
  author   = "Warrington, Shaun and Thompson, Elinor and Bastiani, Matteo and
              Dubois, Jessica and Baxter, Luke and Slater, Rebeccah and Jbabdi,
              Saad and Mars, Rogier B and Sotiropoulos, Stamatios N",
  journal  = "Sci Adv",
  volume   =  8,
  number   =  42,
  pages    = "eabq2022",
  month    =  oct,
  year     =  2022,
  language = "en"
}

@ARTICLE{Schmahmann2010-td,
  title     = "The role of the cerebellum in cognition and emotion: personal
               reflections since 1982 on the dysmetria of thought hypothesis,
               and its historical evolution from theory to therapy",
  author    = "Schmahmann, Jeremy D",
  journal   = "Neuropsychol. Rev.",
  publisher = "Springer Science and Business Media LLC",
  volume    =  20,
  number    =  3,
  pages     = "236--260",
  month     =  sep,
  year      =  2010,
  language  = "en"
}

@ARTICLE{Zeng_undated-iw,
  title     = "Automated Identification of the Retinogeniculate Visual Pathway
               Using a High-Dimensional Tractography Atlas",
  author    = "Zeng, Qingrun and Huang, Jiahao and He, Jianzhong and Chen,
               Shengwei and Xie, Lei and Chen, Zan and Guo, Wenlong and Yao, Sun
               and Li, Mengjun and Li, Mingchu and Feng, Yuanjing",
  journal   = "IEEE Transactions on Cognitive and Developmental Systems",
  publisher = "IEEE",
  volume    = "PP",
  number    =  99,
  pages     = "1--1"
}

@ARTICLE{Sket2019-sx,
  title     = "Neonatal white matter maturation is associated with infant
               language development",
  author    = "Sket, Georgina M and Overfeld, Judith and Styner, Martin and
               Gilmore, John H and Entringer, Sonja and Wadhwa, Pathik D and
               Rasmussen, Jerod M and Buss, Claudia",
  journal   = "Front. Hum. Neurosci.",
  publisher = "Frontiers Media SA",
  volume    =  13,
  pages     =  434,
  month     =  dec,
  year      =  2019,
  language  = "en"
}

@ARTICLE{Wu2021-rf,
  title    = "Highly Reproducible Whole Brain Parcellation in Individuals via
              Voxel Annotation with Fiber Clusters",
  author   = "Wu, Ye and Ahmad, Sahar and Yap, Pew-Thian",
  journal  = "Med. Image Comput. Comput. Assist. Interv.",
  volume   =  12907,
  pages    = "477--486",
  month    =  sep,
  year     =  2021,
  language = "en"
}

@ARTICLE{Khan2024-rq,
  title    = "Sex differences in human brain structure at birth",
  author   = "Khan, Yumnah T and Tsompanidis, Alex and Radecki, Marcin A and
              Dorfschmidt, Lena and {APEX Consortium} and Austin, Topun and
              Suckling, John and Allison, Carrie and Lai, Meng-Chuan and
              Bethlehem, Richard A I and Baron-Cohen, Simon",
  journal  = "Biol. Sex Differ.",
  volume   =  15,
  number   =  1,
  pages    =  81,
  month    =  oct,
  year     =  2024,
  language = "en"
}

@ARTICLE{Vos2013-vn,
  title    = "Multi-fiber tractography visualizations for diffusion {MRI} data",
  author   = "Vos, Sjoerd B and Viergever, Max A and Leemans, Alexander",
  journal  = "PLoS One",
  volume   =  8,
  number   =  11,
  pages    = "e81453",
  month    =  nov,
  year     =  2013,
  language = "en"
}

@ARTICLE{De_Kieviet2012-bv,
  title     = "Brain development of very preterm and very low-birthweight
               children in childhood and adolescence: a meta-analysis",
  author    = "de Kieviet, Jorrit F and Zoetebier, Lydia and van Elburg, Ruurd M
               and Vermeulen, R Jeroen and Oosterlaan, Jaap",
  journal   = "Dev. Med. Child Neurol.",
  publisher = "Wiley",
  volume    =  54,
  number    =  4,
  pages     = "313--323",
  month     =  apr,
  year      =  2012,
  language  = "en"
}

@ARTICLE{Manto2012-yh,
  title     = "Consensus paper: roles of the cerebellum in motor control--the
               diversity of ideas on cerebellar involvement in movement",
  author    = "Manto, Mario and Bower, James M and Conforto, Adriana Bastos and
               Delgado-García, José M and da Guarda, Suzete Nascimento Farias
               and Gerwig, Marcus and Habas, Christophe and Hagura, Nobuhiro and
               Ivry, Richard B and Mariën, Peter and Molinari, Marco and Naito,
               Eiichi and Nowak, Dennis A and Oulad Ben Taib, Nordeyn and
               Pelisson, Denis and Tesche, Claudia D and Tilikete, Caroline and
               Timmann, Dagmar",
  journal   = "Cerebellum",
  publisher = "Springer Science and Business Media LLC",
  volume    =  11,
  number    =  2,
  pages     = "457--487",
  month     =  jun,
  year      =  2012,
  language  = "en"
}

@ARTICLE{Lee2013-km,
  title    = "Radiologic differences in white matter maturation between preterm
              and full-term infants: {TBSS} study",
  author   = "Lee, Ah Young and Jang, Sung Ho and Lee, Eunsil and Ahn, Sang Ho
              and Cho, Hee Kyung and Jo, Hae Min and Son, Su Min",
  journal  = "Pediatr. Radiol.",
  volume   =  43,
  number   =  5,
  pages    = "612--619",
  month    =  mar,
  year     =  2013,
  language = "en"
}

@ARTICLE{Benagiano2018-jz,
  title     = "The functional anatomy of the cerebrocerebellar circuit: A review
               and new concepts",
  author    = "Benagiano, Vincenzo and Rizzi, Anna and Lorusso, Loredana and
               Flace, Paolo and Saccia, Matteo and Cagiano, Raffaele and
               Ribatti, Domenico and Roncali, Luisa and Ambrosi, Glauco",
  journal   = "J. Comp. Neurol.",
  publisher = "Wiley",
  volume    =  526,
  number    =  5,
  pages     = "769--789",
  month     =  apr,
  year      =  2018,
  language  = "en"
}

@ARTICLE{Deoni2012-yj,
  title   = "Investigating white matter development in infancy and early
             childhood using myelin water faction and relaxation time mapping",
  author  = "Deoni, S C and C., Dean Iii D and O'Muircheartaigh, J and
             Dirks, H and Jerskey, B A",
  journal = "Neuroimage",
  volume  =  63,
  number  =  3,
  pages   = "1038--1053",
  year    =  2012
}

@ARTICLE{Zhang2020-qj,
  title    = "Deep white matter analysis ({DeepWMA}): Fast and consistent
              tractography segmentation",
  author   = "Zhang, Fan and Cetin Karayumak, Suheyla and Hoffmann, Nico and
              Rathi, Yogesh and Golby, Alexandra J and O'Donnell, Lauren J",
  journal  = "Med. Image Anal.",
  volume   =  65,
  pages    =  101761,
  month    =  oct,
  year     =  2020,
  language = "en"
}

@ARTICLE{Dimitrova2021-hg,
  title     = "Phenotyping the Preterm Brain: Characterizing Individual
               Deviations From Normative Volumetric Development in Two Large
               Infant Cohorts",
  author    = "Dimitrova, Ralica and Arulkumaran, Sophie and Carney, Olivia and
               Chew, Andrew and Falconer, Shona and Ciarrusta, Judit and
               Wolfers, Thomas and Batalle, Dafnis and Cordero-Grande, Lucilio
               and Price, Anthony N and Teixeira, Rui P A G and Hughes, Emer and
               Egloff, Alexia and Hutter, Jana and Makropoulos, Antonios and
               Robinson, Emma C and Schuh, Andreas and Vecchiato, Katy and
               Steinweg, Johannes K and Macleod, Russell and Marquand, Andre F
               and McAlonan, Grainne and Rutherford, Mary A and Counsell, Serena
               J and Smith, Stephen M and Rueckert, Daniel and Hajnal, Joseph V
               and O'Muircheartaigh, Jonathan and Edwards, A David",
  journal   = "Cereb. Cortex",
  publisher = "academic.oup.com",
  volume    =  31,
  number    =  8,
  pages     = "3665--3677",
  month     =  jul,
  year      =  2021,
  language  = "en"
}

@ARTICLE{Mazziotta2001-tl,
  title    = "A probabilistic atlas and reference system for the human brain:
              International Consortium for Brain Mapping ({ICBM})",
  author   = "Mazziotta, J and Toga, A and Evans, A and Fox, P and Lancaster, J
              and Zilles, K and Woods, R and Paus, T and Simpson, G and Pike, B
              and Holmes, C and Collins, L and Thompson, P and MacDonald, D and
              Iacoboni, M and Schormann, T and Amunts, K and Palomero-Gallagher,
              N and Geyer, S and Parsons, L and Narr, K and Kabani, N and Le
              Goualher, G and Boomsma, D and Cannon, T and Kawashima, R and
              Mazoyer, B",
  journal  = "Philos. Trans. R. Soc. Lond. B Biol. Sci.",
  volume   =  356,
  number   =  1412,
  pages    = "1293--1322",
  month    =  aug,
  year     =  2001,
  language = "en"
}

@article{zhang2020slicerdmri,
  title={SlicerDMRI: diffusion MRI and tractography research software for brain cancer surgery planning and visualization},
  author={Zhang, Fan and Noh, Thomas and Juvekar, Parikshit and Frisken, Sarah F and Rigolo, Laura and Norton, Isaiah and Kapur, Tina and Pujol, Sonia and Wells III, William and Yarmarkovich, Alex and others},
  journal={JCO clinical cancer informatics},
  volume={4},
  pages={299--309},
  year={2020},
  publisher={American Society of Clinical Oncology}
}

@article{norton2017slicerdmri,
  title={SlicerDMRI: open source diffusion MRI software for brain cancer research},
  author={Norton, Isaiah and Essayed, Walid Ibn and Zhang, Fan and Pujol, Sonia and Yarmarkovich, Alex and Golby, Alexandra J and Kindlmann, Gordon and Wassermann, Demian and Estepar, Raul San Jose and Rathi, Yogesh and others},
  journal={Cancer research},
  volume={77},
  number={21},
  pages={e101--e103},
  year={2017},
  publisher={American Association for Cancer Research}
}

@article{dubois2021mri,
  title={MRI of the neonatal brain: a review of methodological challenges and neuroscientific advances},
  author={Dubois, Jessica and Alison, Marianne and Counsell, Serena J and Hertz-Pannier, Lucie and H{\"u}ppi, Petra S and Benders, Manon JNL},
  journal={Journal of Magnetic Resonance Imaging},
  volume={53},
  number={5},
  pages={1318--1343},
  year={2021},
  publisher={Wiley Online Library}
}

@ARTICLE{Pannek2014-ue,
  title    = "Magnetic resonance diffusion tractography of the preterm infant
              brain: a systematic review",
  author   = "Pannek, Kerstin and Scheck, Simon M and Colditz, Paul B and Boyd,
              Roslyn N and Rose, Stephen E",
  journal  = "Dev. Med. Child Neurol.",
  volume   =  56,
  number   =  2,
  pages    = "113--124",
  month    =  feb,
  year     =  2014,
  language = "en"
}

@article{ohuma2023national,
  title={National, regional, and global estimates of preterm birth in 2020, with trends from 2010: a systematic analysis},
  author={Ohuma, Eric O and Moller, Ann-Beth and Bradley, Ellen and Chakwera, Samuel and Hussain-Alkhateeb, Laith and Lewin, Alexandra and Okwaraji, Yemisrach B and Mahanani, Wahyu Retno and Johansson, Emily White and Lavin, Tina and others},
  journal={The Lancet},
  volume={402},
  number={10409},
  pages={1261--1271},
  year={2023},
  publisher={Elsevier}
}

@article{perin2022global,
  title={Global, regional, and national causes of under-5 mortality in 2000--19: an updated systematic analysis with implications for the Sustainable Development Goals},
  author={Perin, Jamie and Mulick, Amy and Yeung, Diana and Villavicencio, Francisco and Lopez, Gerard and Strong, Kathleen L and Prieto-Merino, David and Cousens, Simon and Black, Robert E and Liu, Li},
  journal={The Lancet Child \& Adolescent Health},
  volume={6},
  number={2},
  pages={106--115},
  year={2022},
  publisher={Elsevier}
}

@ARTICLE{Elam2021-gc,
  title     = "The Human Connectome Project: A retrospective",
  author    = "Elam, Jennifer Stine and Glasser, Matthew F and Harms, Michael P
               and Sotiropoulos, Stamatios N and Andersson, Jesper L R and
               Burgess, Gregory C and Curtiss, Sandra W and Oostenveld, Robert
               and Larson-Prior, Linda J and Schoffelen, Jan-Mathijs and Hodge,
               Michael R and Cler, Eileen A and Marcus, Daniel M and Barch,
               Deanna M and Yacoub, Essa and Smith, Stephen M and Ugurbil, Kamil
               and Van Essen, David C",
  journal   = "Neuroimage",
  publisher = "Elsevier",
  volume    =  244,
  pages     =  118543,
  month     =  dec,
  year      =  2021,
  language  = "en"
}

@article{brown2011brain,
  title={Brain network local interconnectivity loss in aging APOE-4 allele carriers},
  author={Brown, Jesse A and Terashima, Kevin H and Burggren, Alison C and Ercoli, Linda M and Miller, Karen J and Small, Gary W and Bookheimer, Susan Y},
  journal={Proceedings of the National Academy of Sciences},
  volume={108},
  number={51},
  pages={20760--20765},
  year={2011},
  publisher={National Academy of Sciences}
}

@article{drakesmith2015overcoming,
  title={Overcoming the effects of false positives and threshold bias in graph theoretical analyses of neuroimaging data},
  author={Drakesmith, Mark and Caeyenberghs, Karen and Dutt, Anirban and Lewis, Glyn and David, Anthony S and Jones, Derek K},
  journal={Neuroimage},
  volume={118},
  pages={313--333},
  year={2015},
  publisher={Elsevier}
}

@article{shany2017diffusion,
  title={Diffusion tensor tractography of the cerebellar peduncles in prematurely born 7-year-old children},
  author={Shany, Eilon and Inder, Terrie E and Goshen, Sharon and Lee, Iris and Neil, Jeffrey J and Smyser, Christopher D and Doyle, Lex W and Anderson, Peter J and Shimony, Joshua S},
  journal={The Cerebellum},
  volume={16},
  number={2},
  pages={314--325},
  year={2017},
  publisher={Springer}
}

@article{kimpton2021diffusion,
  title={Diffusion magnetic resonance imaging assessment of regional white matter maturation in preterm neonates},
  author={Kimpton, JA and Batalle, Dafnis and Barnett, ML and Hughes, EJ and Chew, ATM and Falconer, Shona and Tournier, Jacques-Donald and Alexander, D and Zhang, Hui and Edwards, AD and others},
  journal={Neuroradiology},
  volume={63},
  number={4},
  pages={573--583},
  year={2021},
  publisher={Springer}
}

@article{gimenez2008accelerated,
  title={Accelerated cerebral white matter development in preterm infants: a voxel-based morphometry study with diffusion tensor MR imaging},
  author={Gim{\'e}nez, M{\'o}nica and Miranda, Maria J and Born, A Peter and Nagy, Zoltan and Rostrup, Egill and Jernigan, Terry L},
  journal={Neuroimage},
  volume={41},
  number={3},
  pages={728--734},
  year={2008},
  publisher={Elsevier}
}

@article{berman2005quantitative,
  title={Quantitative diffusion tensor MRI fiber tractography of sensorimotor white matter development in premature infants},
  author={Berman, Jeffrey I and Mukherjee, Pratik and Partridge, Savannah C and Miller, Steven P and Ferriero, Donna M and Barkovich, A James and Vigneron, Daniel B and Henry, Roland G},
  journal={Neuroimage},
  volume={27},
  number={4},
  pages={862--871},
  year={2005},
  publisher={Elsevier}
}

\clearpage
\section*{Supplement}
\appendix
\renewcommand{\thefigure}{S\arabic{figure}}
\renewcommand{\thetable}{S\arabic{table}}
\setcounter{figure}{0}
\setcounter{table}{0}

{\small
\begin{longtable}[h]{
>{\centering\arraybackslash}p{2cm}|%
>{\centering\arraybackslash}p{1.5cm}%
>{\centering\arraybackslash}p{1.6cm}|%
>{\centering\arraybackslash}p{1.5cm}%
>{\centering\arraybackslash}p{1.5cm}|%
>{\centering\arraybackslash}p{1.5cm}%
>{\centering\arraybackslash}p{1.5cm}}
\caption{Summary of FA beta coefficients across populations. 
The table lists the beta coefficient values corresponding to the comparisons shown in 
Fig.~\ref{dHCP HCP beta}, Fig.~\ref{male female beta}, and Fig.~\ref{fullterm preterm beta}.}
\label{tab:beta-fa-matrix}\\
\hline
\textbf{Tract} & \textbf{dHCP} & \textbf{HCP--YA} & \textbf{Full-term} & \textbf{Preterm} & \textbf{Male} & \textbf{Female} \\
\hline
\endfirsthead

\hline
\textbf{Tract} & \textbf{dHCP} & \textbf{HCP--YA} & \textbf{Full-term} & \textbf{Preterm} & \textbf{Male} & \textbf{Female} \\
\hline
\endhead

\hline
\multicolumn{7}{r}{\small Continued on next page}
\\\endfoot

\hline
\endlastfoot
AF\_left & 0.022820 & -0.000168 & 0.022266 & 0.013753 & 0.018338 & 0.024170 \\
AF\_right & 0.019116 & -0.000698 & 0.018217 & 0.013365 & 0.019496 & 0.016842 \\
EC\_left & 0.014653 & 0.000489 & 0.014681 & 0.010020 & 0.012136 & 0.015086 \\
EC\_right & 0.013918 & 0.000576 & 0.013672 & 0.009805 & 0.013602 & 0.013664 \\
EmC\_left & 0.017854 & 0.000208 & 0.018151 & 0.012162 & 0.015257 & 0.019197 \\
EmC\_right & 0.016455 & 0.000121 & 0.016794 & 0.010979 & 0.015268 & 0.018567 \\
ILF\_left & 0.019183 & 0.000200 & 0.019554 & 0.012668 & 0.015802 & 0.021859 \\
ILF\_right & 0.018033 & -0.000214 & 0.017877 & 0.012041 & 0.014234 & 0.019699 \\
IOFF\_left & 0.019137 & 0.000283 & 0.019669 & 0.013623 & 0.016278 & 0.021794 \\
IOFF\_right & 0.017648 & -0.000014 & 0.018280 & 0.015094 & 0.016781 & 0.019278 \\
MdLF\_left & 0.014732 & 0.000291 & 0.015245 & 0.009625 & 0.012930 & 0.017442 \\
MdLF\_right & 0.013680 & 0.000046 & 0.013839 & 0.009744 & 0.010414 & 0.016073 \\
SLF-I\_left & 0.015346 & -0.000213 & 0.015077 & 0.011669 & 0.012872 & 0.014961 \\
SLF-I\_right & 0.015896 & -0.000323 & 0.015841 & 0.012076 & 0.015437 & 0.014774 \\
SLF-II\_left & 0.019734 & -0.000116 & 0.020024 & 0.012469 & 0.017591 & 0.021266 \\
SLF-II\_right & 0.019175 & -0.000296 & 0.019575 & 0.010881 & 0.018980 & 0.020407 \\
SLF-III\_left & 0.017150 & -0.000107 & 0.017358 & 0.015848 & 0.014939 & 0.018797 \\
SLF-III\_right & 0.018007 & -0.000061 & 0.018129 & 0.014309 & 0.016841 & 0.018644 \\
UF\_left & 0.016463 & -0.000674 & 0.016522 & 0.009400 & 0.013880 & 0.017540 \\
UF\_right & 0.014901 & -0.000680 & 0.015431 & 0.008215 & 0.014614 & 0.016258 \\
CC1 & 0.016317 & -0.001371 & 0.016475 & 0.009437 & 0.014635 & 0.017262 \\
CC2 & 0.017053 & -0.001554 & 0.017395 & 0.011229 & 0.016904 & 0.018147 \\
CC3 & 0.017183 & -0.000642 & 0.016732 & 0.013636 & 0.017096 & 0.017872 \\
CC4 & 0.017275 & -0.000358 & 0.016719 & 0.018019 & 0.015145 & 0.018384 \\
CC5 & 0.017199 & -0.000461 & 0.016319 & 0.017016 & 0.016547 & 0.017311 \\
CC6 & 0.010845 & -0.000617 & 0.010730 & 0.008775 & 0.009524 & 0.011308 \\
CC7 & 0.009407 & -0.000408 & 0.009547 & 0.009754 & 0.009051 & 0.008574 \\
CB\_D\_left & 0.010623 & -0.000166 & 0.010206 & 0.007990 & 0.008674 & 0.011235 \\
CB\_D\_right & 0.010578 & 0.000277 & 0.010911 & 0.007656 & 0.011006 & 0.011575 \\
CB\_V\_left & 0.010774 & 0.000347 & 0.010997 & 0.009121 & 0.009433 & 0.012588 \\
CB\_V\_right & 0.011054 & 0.000263 & 0.011156 & 0.010260 & 0.011080 & 0.012870 \\
CST\_left & 0.010159 & 0.000219 & 0.010224 & 0.015698 & 0.009398 & 0.010658 \\
CST\_right & 0.010802 & 0.000145 & 0.010744 & 0.014560 & 0.010120 & 0.011482 \\
CR-F\_left & 0.012559 & 0.000063 & 0.012662 & 0.011975 & 0.012597 & 0.013782 \\
CR-F\_right & 0.012889 & -0.000079 & 0.012959 & 0.012864 & 0.012882 & 0.013310 \\
CR-P\_left & 0.007810 & 0.000036 & 0.007878 & 0.011814 & 0.006012 & 0.008887 \\
CR-P\_right & 0.008877 & -0.000113 & 0.008602 & 0.011158 & 0.007718 & 0.009107 \\
SF\_left & 0.010702 & -0.000112 & 0.009612 & 0.007806 & 0.010259 & 0.009536 \\
SF\_right & 0.008862 & -0.000352 & 0.008112 & 0.008092 & 0.008585 & 0.006819 \\
SO\_left & 0.017845 & 0.000011 & 0.017816 & 0.014089 & 0.015062 & 0.019666 \\
SO\_right & 0.016159 & 0.000002 & 0.016227 & 0.014077 & 0.014800 & 0.017610 \\
SP\_left & 0.015851 & 0.000172 & 0.015087 & 0.013395 & 0.015697 & 0.014201 \\
SP\_right & 0.013935 & 0.000197 & 0.013158 & 0.011815 & 0.013420 & 0.014015 \\
TF\_left & 0.013558 & 0.000082 & 0.013438 & 0.010075 & 0.013247 & 0.014698 \\
TF\_right & 0.012391 & -0.000133 & 0.012518 & 0.008869 & 0.012424 & 0.013680 \\
TO\_left & 0.013670 & -0.000342 & 0.011503 & 0.011374 & 0.007685 & 0.012790 \\
TO\_right & 0.013101 & -0.000587 & 0.009743 & 0.010452 & 0.006004 & 0.010012 \\
TT\_left & 0.009707 & 0.000333 & 0.008555 & 0.006620 & 0.005910 & 0.011353 \\
TT\_right & 0.009445 & 0.000042 & 0.008294 & 0.008896 & 0.007551 & 0.009525 \\
TP\_left & 0.010925 & 0.000084 & 0.011048 & 0.013515 & 0.010999 & 0.011923 \\
TP\_right & 0.010622 & 0.000024 & 0.011157 & 0.012924 & 0.011050 & 0.011905 \\
CPC\_left & 0.008830 & -0.000247 & 0.008701 & 0.012811 & 0.007166 & 0.009107 \\
CPC\_right & 0.009294 & 0.000112 & 0.008619 & 0.012363 & 0.008659 & 0.007681 \\
ICP\_left & 0.006841 & 0.000591 & 0.006208 & 0.011849 & 0.005394 & 0.007268 \\
ICP\_right & 0.005900 & 0.000352 & 0.005478 & 0.009419 & 0.002333 & 0.005061 \\
Intra-CBLM-I\&P\_left & 0.001967 & 0.000090 & 0.002468 & 0.001580 & 0.000981 & 0.002926 \\
Intra-CBLM-I\&P\_right & 0.001427 & -0.000110 & 0.001372 & 0.002163 & 0.001371 & 0.000875 \\
Intra-CBLM-PaT\_left & 0.003239 & -0.000296 & 0.003336 & 0.003040 & 0.003735 & 0.003448 \\
Intra-CBLM-PaT\_right & 0.002925 & -0.000440 & 0.002948 & 0.003714 & 0.003314 & 0.002712 \\
MCP & 0.004264 & -0.000025 & 0.006119 & 0.013596 & 0.007868 & 0.006595 \\
SCP\_left & 0.008263 & 0.001363 & 0.008835 & 0.012352 & 0.007988 & 0.007836 \\
SCP\_right & 0.007678 & 0.001009 & 0.007775 & 0.011963 & 0.007641 & 0.008790 \\
Sup-F\_left & 0.016633 & -0.000521 & 0.016366 & 0.012263 & 0.014698 & 0.017330 \\
Sup-F\_right & 0.016988 & -0.000578 & 0.016808 & 0.012000 & 0.015652 & 0.016862 \\
Sup-FP\_left & 0.014341 & -0.000088 & 0.014603 & 0.013161 & 0.014757 & 0.013613 \\
Sup-FP\_right & 0.015379 & 0.000068 & 0.014972 & 0.012617 & 0.015310 & 0.013664 \\
Sup-O\_left & 0.010292 & -0.000369 & 0.009762 & 0.004157 & 0.008067 & 0.010705 \\
Sup-O\_right & 0.008949 & -0.000300 & 0.008662 & 0.004365 & 0.005775 & 0.009150 \\
Sup-OT\_left & 0.018054 & 0.000251 & 0.017326 & 0.011693 & 0.012212 & 0.018288 \\
Sup-OT\_right & 0.017288 & -0.000357 & 0.016762 & 0.016590 & 0.011633 & 0.017245 \\
Sup-P\_left & 0.014059 & -0.000410 & 0.014130 & 0.007491 & 0.012779 & 0.014240 \\
Sup-P\_right & 0.013559 & -0.000315 & 0.012540 & 0.007044 & 0.012786 & 0.011191 \\
Sup-PO\_left & 0.014940 & -0.000297 & 0.015600 & 0.009947 & 0.017976 & 0.016624 \\
Sup-PO\_right & 0.012243 & -0.000429 & 0.012265 & 0.010710 & 0.011962 & 0.010477 \\
Sup-PT\_left & 0.014961 & -0.000210 & 0.015256 & 0.008636 & 0.013170 & 0.018523 \\
Sup-PT\_right & 0.014080 & -0.000190 & 0.014165 & 0.010144 & 0.011542 & 0.017349 \\
Sup-T\_left & 0.014249 & -0.000068 & 0.013918 & 0.007911 & 0.011028 & 0.015223 \\
Sup-T\_right & 0.011948 & -0.000360 & 0.011473 & 0.007225 & 0.006785 & 0.014055 \\
\end{longtable}
}

\begin{figure}[h]
  \centering
  \begin{subfigure}[b]{0.85\textwidth}
    \centering
    \includegraphics[width=\textwidth]{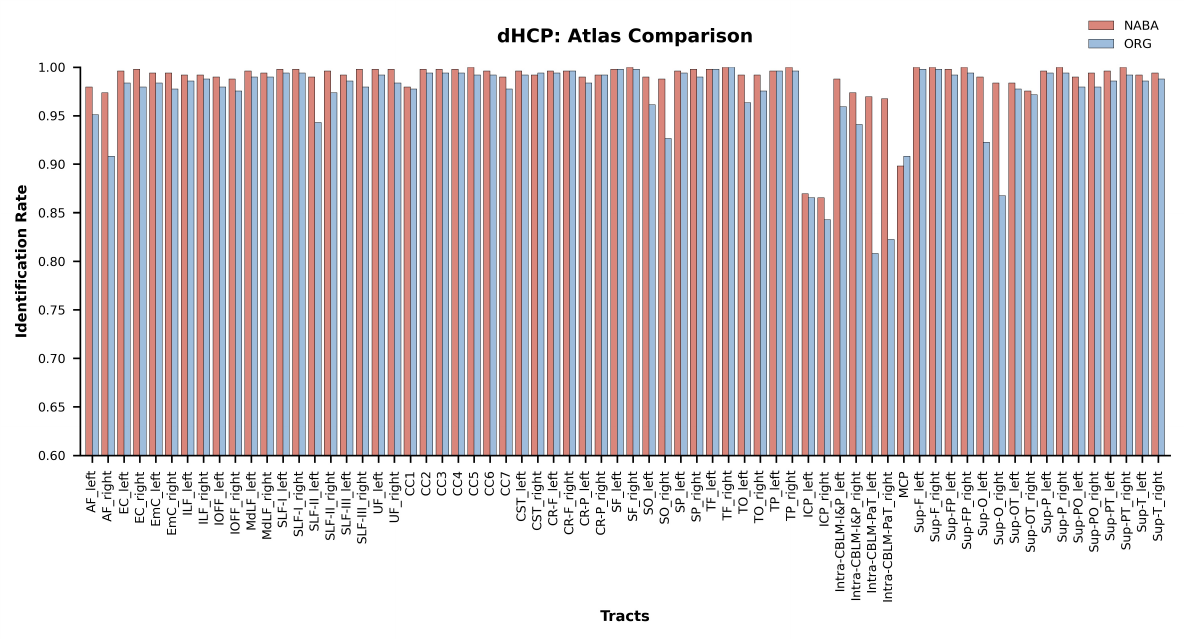}
    \caption{dHCP}
    \label{fig:subfig_a}
  \end{subfigure}
  \hfill
  \begin{subfigure}[b]{0.85\textwidth}
    \centering
    \includegraphics[width=\textwidth]{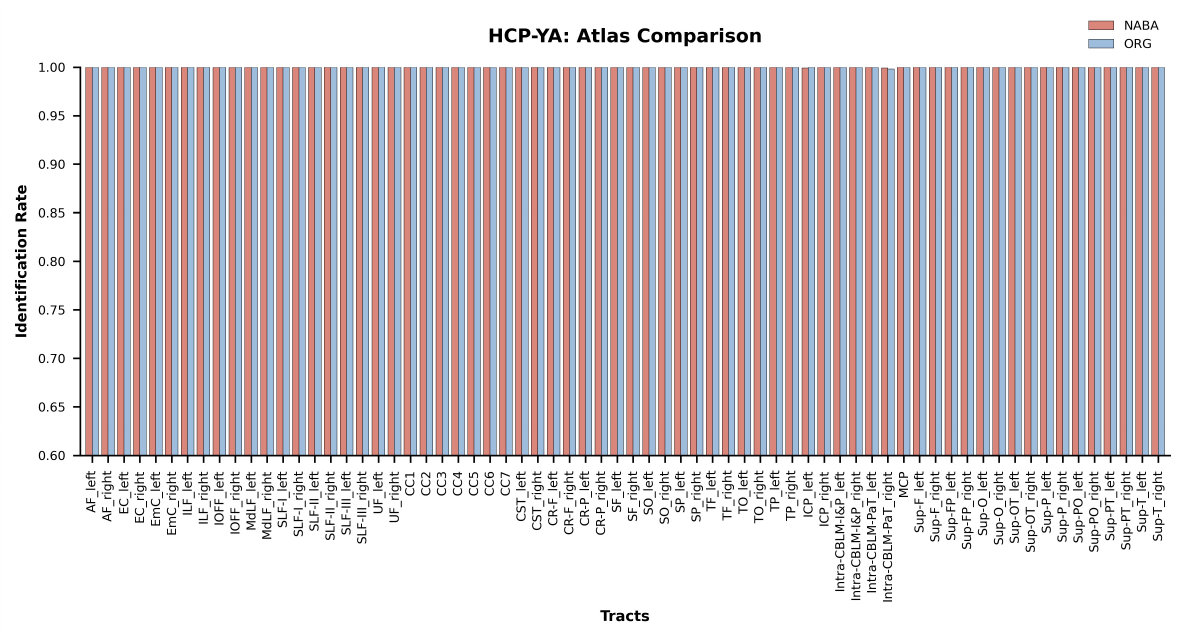}
    \caption{HCP-YA}
    \label{fig:subfig_b}
  \end{subfigure}
  \caption{IR comparison between NABA atlas and ORG atlas on dHCP and HCP-YA (threshold = 5).}
  \label{fig:atlas comparison 5}
\end{figure}

\begin{figure}[h]
  \centering
  \begin{subfigure}[b]{0.85\textwidth}
    \centering
    \includegraphics[width=\textwidth]{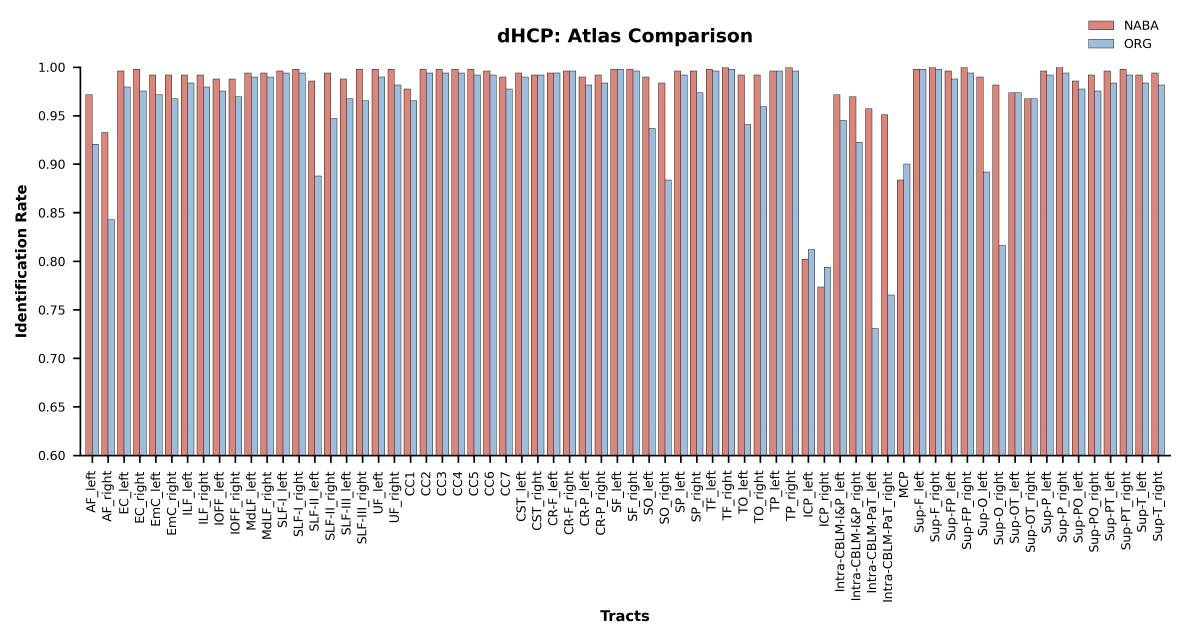}
    \caption{dHCP}
    \label{fig:subfig_a}
  \end{subfigure}
  \hfill
  \begin{subfigure}[b]{0.85\textwidth}
    \centering
    \includegraphics[width=\textwidth]{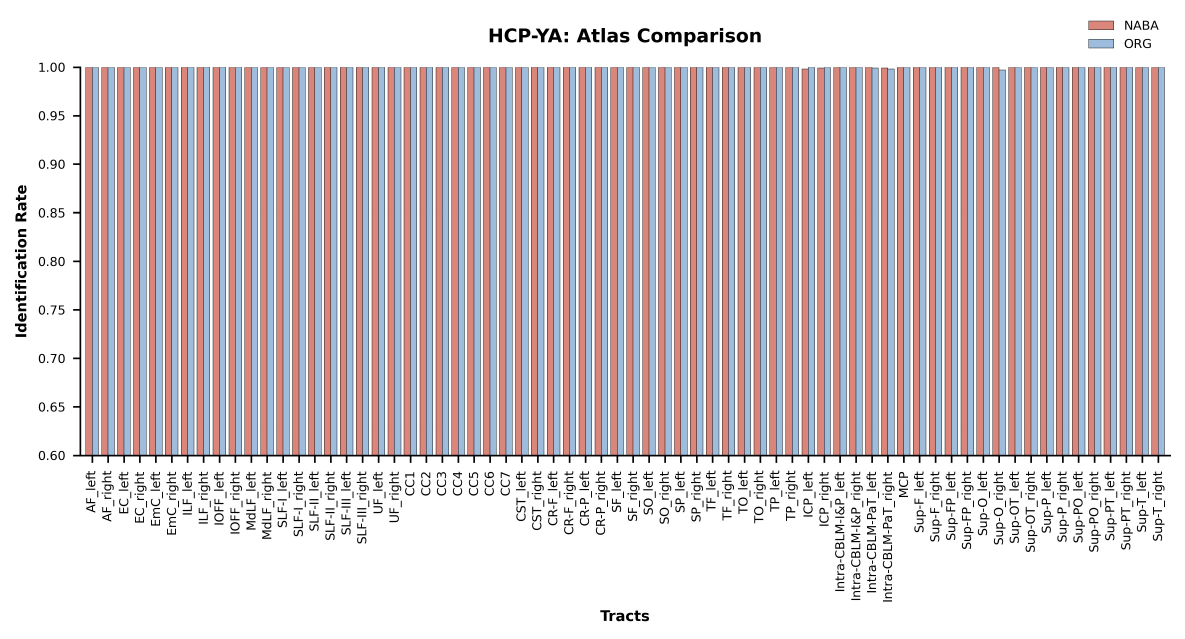}
    \caption{HCP-YA}
    \label{fig:subfig_b}
  \end{subfigure}
  \caption{IR comparison between NABA atlas and ORG atlas on dHCP and HCP-YA (threshold = 15).}
  \label{fig:atlas comparison 15}
\end{figure}

\begin{figure}[h]
    \centering
    \includegraphics[width=1\textwidth]{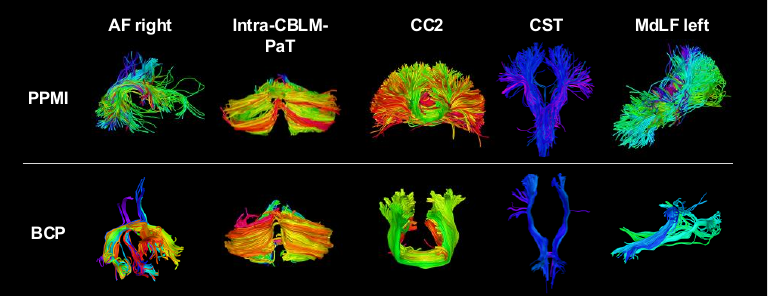}
    \caption{Visualization of NABA atlas parcellation on additional diffusion MRI datasets. A young adult subject from the Parkinson’s Progression Markers Initiative (PPMI) dataset((PPMI) Parkinson’s Progression Marker Initiative 2012): 31 years old; b = 1000 s/mm²; 64 directions; single-shell; 2×2×2 mm resolution. Tractography reconstructed using UKF. A neonate subject from the Baby Connectome Project (BCP) dataset(Howell et al. 2019): 2.8 months; gestational age at birth = 39.3 weeks; two-shell acquisition (37 directions per shell); 1.5×1.5×1.5 mm resolution. Tractography reconstructed using DSI Studio.}
    \label{fig:PPMI BCP}
\end{figure}


\begin{figure}[h]
    \centering
    \includegraphics[width=\textwidth]{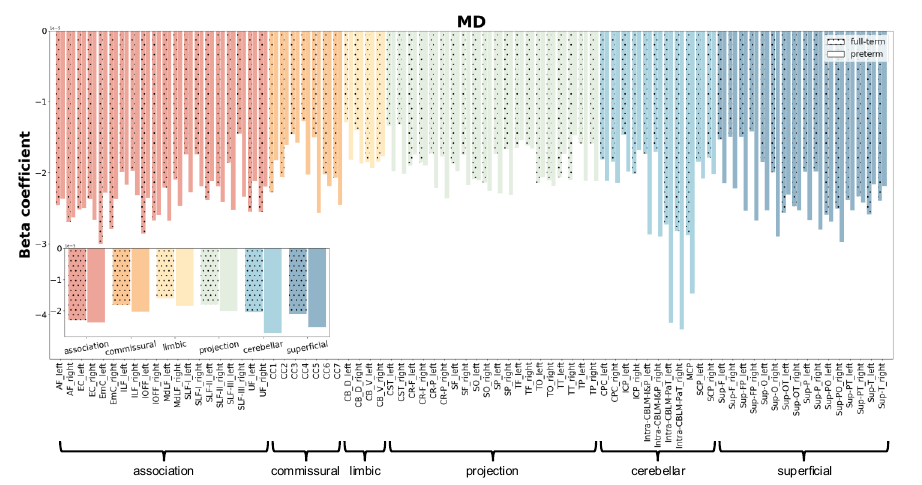}
    \caption{Rate of change with age (beta coefficient) of MD in full-term and preterm neonates in dHCP. The two groups were sex-balanced, and the GLM controlled for birth weight and head circumference at scan. The two groups are distinguished using different marker styles for the bars. Tracts of different categories are color-coded. The upper right corner displays the rate of change at the category level for both populations.}
    \label{fig:fullterm preterm beta MD}
\end{figure}

\begin{figure}[h]
    \centering
    \includegraphics[width=0.7\textwidth]{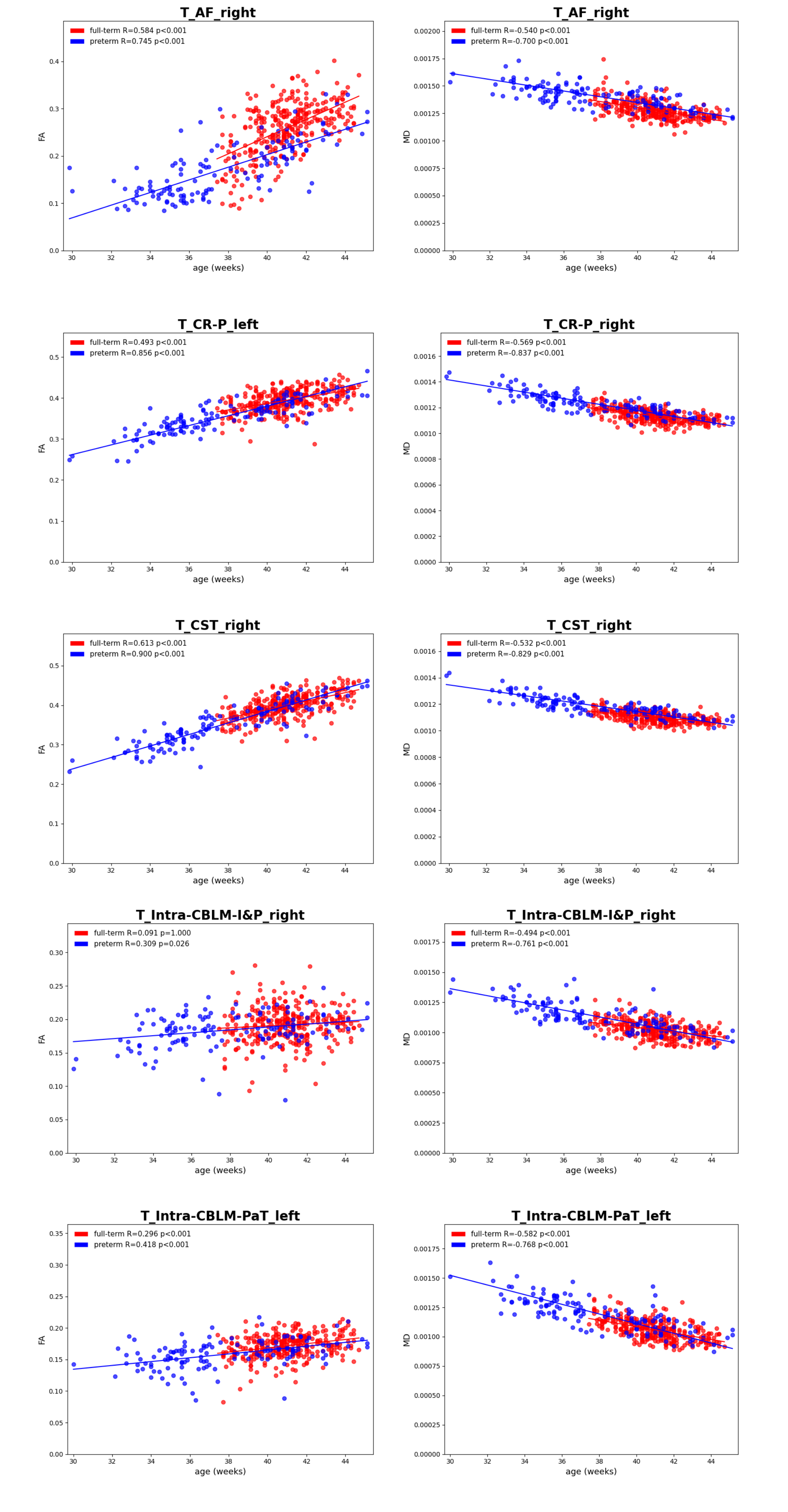}
    \caption{Scatter plots of FA and MD versus scan age for selected WM tracts in full-term and preterm neonates.}
    \label{fig:scatter}
\end{figure}

\end{document}